\documentclass[preprint,authoryear]{elsarticle}
\usepackage{graphicx}
\usepackage{amsfonts}
\usepackage{mathrsfs}
\usepackage{amssymb}
\usepackage{amsmath}
\usepackage{blkarray}
\usepackage[dvipsnames,table]{xcolor}
\newcolumntype{L}{>{\raggedright\arraybackslash}X}
\newcolumntype{C}[1]{>{\centering\let\newline\\\arraybackslash\hspace{0pt}}m{#1}}
\newcolumntype{W}{>{\hspace{-1ex}}}
\newcolumntype{S}{>{\hspace{-2ex}}}
\newcolumntype{Q}{>{\hspace{-2.5ex}}}
\usepackage{textcomp}
\usepackage{subfig}
\usepackage[utf8]{inputenc}
\usepackage[T1]{fontenc}
\usepackage{nccmath}
\usepackage{booktabs}
\usepackage{algorithm2e}
\usepackage{float}
\newtheorem{definition}{Definition}
\usepackage{multirow}
\usepackage{hyperref}

\begin{document}
\begin{frontmatter}
    \title{Encoder-Decoder Model for Suffix Prediction in Predictive Monitoring}

    \author[label1]{Efrén Rama-Maneiro}
    \author[label1]{Pablo Monteagudo-Lago}
    \author[label1,label2]{Juan C. Vidal}
    \author[label1]{Manuel Lama}
%	\affiliation[label1]{
%            organization={Centro Singular de Investigación en Tecnoloxías Intelixentes (CiTIUS)},
%            addressline={Universidade de Santiago de Compostela}, 
%            city={Santiago de Compostela},
%            country={Spain}
%        }
%    \affiliation[label2]{
%            organization={Departamento de Electrónica e Computación},
%            addressline={Universidade de Santiago de Compostela}, 
%            city={Santiago de Compostela},
%            country={Spain}
%    }
	\address[label1]{Centro Singular de Investigación en Tecnoloxías Intelixentes (CiTIUS), Universidade de Santiago de Compostela, Santiago de Compostela, Spain}
	\address[label2]{Departamento de Electrónica e Computación, Universidade de Santiago de Compostela, Santiago de Compostela, Spain}

    %%%%%%%%%%%%%%%%
    \begin{abstract}
    %%%%%%%%%%%%%%%%
    % Predictive monitoring is a subfield of process mining that aims to predict how a running case will unfold in the future. One of its main challenges is forecasting the sequence of activities that will occur from a given point in time. This paper proposes a novel architecture based on both an encoder-decoder model with an attention mechanism and a heuristic search algorithm to predict the sequence of activities that will follow a running case. Our approach has been tested using 12 public event logs against 6 different state-of-the-art solutions, showing that it significantly outperforms these solutions for the suffix prediction in predictive monitoring.
    Predictive monitoring is a subfield of process mining that aims to predict how a running case will unfold in the future. One of its main challenges is forecasting the sequence of activities that will occur from a given point in time ---suffix prediction---. Most approaches to the suffix prediction problem learn to predict the suffix by learning how to predict the next activity only, not learning from the whole suffix during the training phase. This paper proposes a novel architecture based on an encoder-decoder model with an attention mechanism that decouples the representation learning of the prefixes from the inference phase, predicting only the activities of the suffix. During the inference phase, this architecture is extended with a heuristic search algorithm that improves the selection of the activity for each index of the suffix. Our approach has been tested using 12 public event logs against 6 different state-of-the-art proposals, showing that it significantly outperforms these proposals.
    \end{abstract}
    
\end{frontmatter}

%%%%%%%%%%%%%%%%%%%%%%
\section{Introduction}
\label{sec:intro}
%%%%%%%%%%%%%%%%%%%%%%

Process mining has emerged as a discipline that aims to discover, monitor, and improve business processes by extracting knowledge from past execution data that are recorded in an event log~\citep{van_der_aalst_process_2016}. One of the process mining tasks that has received a lot of attention in the past years is forecasting how a case is going to unfold. This allows organizations to prevent undesired outcomes, issues, or delays~\citep{francescomarino_2022}. This task belongs to a subfield of process mining, known as \textit{predictive monitoring}, which leverages machine learning techniques in order to learn a predictive model over the event log. Many predictive problems can be defined, such as forecasting the next activity, the next timestamp, the remaining time until the end of the case, or the next sequence of activities. This paper focuses on the latter problem, aka activity suffix prediction.

Different machine learning techniques have been used in predictive monitoring~\citep{Francescomarino2019, Lee2018, Leontjeva2015, Leoni2016}, but deep learning still holds the best overall results~\citep{tax_interdisciplinary_2020}. Among these latter techniques, recurrent neural networks (RNNs) are one of the most used models due to their adequacy to the sequential nature of business process logs. However, other types of deep neural networks, such as Convolutional Neural Networks (CNNs)~\citep{pasquadibisceglie_cnn_2019, Mauro2019}, Graph Neural Networks (GNNs)~\citep{Venugopal2021, rama-maneiro_embedding_2021}, Generative Adversarial Networks (GANs)~\citep{adversarial_taymouri_2021}, or autoencoders~\citep{Mehdiyev2018} have also achieved competitive results. Regarding the suffix prediction problem, several deep learning approaches have been proposed, either based on encoder-decoder architectures~\citep{adversarial_taymouri_2021, averages_ketyko_2022, attention_jalayer_2020, mann_khan_2021}, or not~\citep{Tax2017, Evermann2017, Francescomarino2017, weinzierl_prescriptive_2020, camargo_learning_2019, entity_dalmas_2021, remaining_sun_2021, averages_ketyko_2022, mmpred_lin_2019, robust_venkateswaran_2021, reinforcement_agarwal_2022}.

%On the one hand, the approaches that are not based on encoder-decoder architectures are forced to predict the same features they use in their input. In this case, either they only use the activities as inputs~\citep{Evermann2017}, losing information that is available in the event log, or they also use additional information, which accumulates errors during the inference phase by predicting features that are not relevant to the predictive task. One way to avoid this problem could be to group the whole prefixes and suffixes into categories through embeddings, but this process is bound to lose information due to the grouping process itself. On the other hand, approaches based on encoder-decoder architectures do not truly exploit the capabilities of these types of neural models with attention mechanisms. The encoder and decoder of these models are trained to predict the same features so they still have the aforementioned issues, with the exception of~\citep{attention_jalayer_2020}, approach which does not learn to predict the suffix but the next activity, so it does not have information from the whole target suffix during the training phase. In the inference phase, all approaches ---except~\citep{Francescomarino2017} and \citep{adversarial_taymouri_2021}--- rely on selecting the most probable activity or on randomly selecting this activity based on activities probability distribution in the prediction.

On the one hand, approaches based on encoder-decoder architectures do not truly exploit the capabilities of these types of neural models with attention mechanisms. The encoder and decoder of these models are trained to predict the same features so they still have the aforementioned issues, with the exception of~\citep{attention_jalayer_2020} approach, which does not learn to predict the suffix but the next activity, so it does not have information from the whole target suffix during the training phase. In the inference phase, all approaches ---except~\citep{Francescomarino2017} and \citep{adversarial_taymouri_2021}--- rely on selecting the most probable activity or on randomly selecting this activity based on the activities probability distribution in the prediction. On the other hand, the approaches that are not based on encoder-decoder architectures are forced to predict the same features they use in their input. In this case, either they only use the activities as inputs~\citep{Evermann2017}, losing information that is available in the event log, or they also use additional information, which accumulates errors during the inference phase by predicting features that are not relevant to the predictive task. One way to avoid this problem could be to group the whole prefixes and suffixes into categories through embeddings, but this process is bound to lose information due to the grouping process itself.

In this paper, we present ASTON, an architecture based on an encoder-decoder with an attention mechanism that helps to retain information when the input prefixes are considerably long. Furthermore, we argue that the strategy used to select the next activity during the suffix prediction phase is essential to obtain a good performance. In particular, we propose the usage of a beam search strategy because it brings a balance between a greedy search and exhaustively exploring the whole tree of possible suffixes. Unlike~\citep{Francescomarino2017}, we also state that length normalization during the beam search is essential to avoid predicting suffixes that are too short. In summary, the main contributions of our approach are as follows:

\begin{itemize}
    \item An encoder-decoder neural model with an attention mechanism that is trained to predict the activity suffix, instead of being trained to predict the next activity. Our neural model decouples the prediction of the suffix from the representation learning of the prefix, so it does not need to learn to predict features irrelevant to the prediction task.
    
    \item A beam search strategy that selects the most promising suffix based on a heuristic search, achieving a good balance between the approach performance and the computational time to obtain a result.
    
    \item A validation of ASTON using 12 real-life event logs, comparing the results against six different approaches of the state of the art. These results show that ASTON obtains the best results and that there are statistically significant differences with the other approaches.
\end{itemize}

The remainder of this paper is structured as follows. Section \ref{sec:related-work} presents the state-of-the-art predictive monitoring techniques based on deep learning; Section \ref{sec:preliminaries} introduces formally the definitions and concepts that are necessary to understand the approach; Section \ref{sec:abasp} describes our approach in detail; Section \ref{sec:evaluation} shows the evaluation of our approach using real-life event logs, and Section \ref{sec:conclusions} highlights the conclusions of the paper and future work.

%%%%%%%%%%%%%%%%%%%%%%%%
\section{Related work}
\label{sec:related-work}
%%%%%%%%%%%%%%%%%%%%%%%%

The first approaches that focused on predicting the activity suffix of a running case using deep learning were~\citep{Tax2017} and~\citep{Evermann2017}. \citep{Tax2017} uses an LSTM neural network to predict the next activity and the time until the next event, encoding the activities using one-hot encoding and adding multiple time features extracted from the events. Then, during the suffix prediction phase, both the time and the next activity are iteratively predicted, which penalizes the overall activity suffix prediction since errors between successive event predictions are accumulated until the end. In~\citep{Evermann2017} a similar approach is followed, but it only uses the activities of the event log, which are encoded as embeddings, meaning that relevant information for the activity suffix prediction, such as time- and resource-related features, is missing. Furthermore, both approaches use either argmax sampling~\citep{Tax2017} or random sampling~\citep{Evermann2017}, which does not select the suffix with the highest probability overall.

\citep{Francescomarino2017} is built upon the architecture of~\citep{Tax2017} but it controls the selection of the next activity during the suffix prediction phase by checking whether the prediction is compliant or not with a set of LTL-mined rules. It uses a beam search to generate the suffix prediction. \citep{weinzierl_prescriptive_2020} is also built upon the work of~\citep{Tax2017}, but it focuses on recommending the best action ---most appropriate activity according to a key process indicator, KPI--- for a running case, predicting the suffix as a support for their system. It trains a multitask LSTM network to predict both the next activity and the KPI for a given event. Then, it predicts the suffix through the neural network and selects the most similar suffix that is compliant with a process model and maximizes the target KPI. Using that suffix, the next best action is recommended. These approaches suffer from the same drawbacks as~\citep{Tax2017}. Furthermore,~\citep{Francescomarino2017} does not normalize the beam search so this approach tends to predict suffixes that are shorter than the ground truth ones.

Other approaches focus on finding novel ways to encode the information from the event log. \citep{camargo_learning_2019} clusters the resources into roles and then learns embeddings for both activities and roles. It iteratively predicts each activity of the suffix from a prefix by using an LSTM that is trained to predict both the next activity as well as the role that performs the activity and the time to the next event. \citep{entity_dalmas_2021} encodes the whole prefix into a single representation through embeddings as well as some temporal information such as the time between events, the hour and the day of the month for each event, etc. Then, a feed-forward network is trained to predict the next suffix identifier, hence achieving a one-shot suffix prediction. \citep{remaining_sun_2021} follows a similar approach to~\citep{weinzierl_prescriptive_2020}. First, it replays the prefixes on a Petri net in order to obtain the set of activated places ---behavioral context---; then, an LSTM is trained to predict the next sequence of attributes given a prefix of attributes ---data context---; and, finally, given a certain prefix during inference, the most similar suffix from the training set is selected according to both its behavioral context and data context. These approaches still suffer the same drawbacks as~\citep{Tax2017}, which are even aggravated since they use more information than~\citep{Tax2017}, so more errors are accumulated in the inference phase. Furthermore, in~\citep{entity_dalmas_2021} the encoding step of both whole prefixes and suffixes into a single embedding is prone to lose a lot of information, so the predictions are bound to be inaccurate. \citep{robust_venkateswaran_2021} proposes to split the prefixes into buckets and then search for a set of input attributes of the event log whose prediction probability is consistent across every set of attributes ---invariant set of attributes---. This search procedure is done by learning an LSTM with a technique called Invariant Risk Minimization \citep{risk_arjovsky_2019}.

Some approaches are based on novel architectures to deal with the suffix prediction problem. \citep{adversarial_taymouri_2021} presents an encoder-decoder LSTM that is trained as a Generative Adversarial Network ---GAN---. Specifically, the encoder-decoder acts as a generator of suffixes that aims to fool the discriminator, which is also an LSTM trained to distinguish between fake suffixes from the generator and real suffixes. \citep{averages_ketyko_2022} compares multiple architectures for suffix prediction such as LSTMs, recurrent autoencoders ---either trained by maximum likelihood estimation or by using a GAN approach---, or transformers~\citep{vaswani_attention_2017}, concluding that no architecture clearly outperforms the other ones in a wide variety of event logs. \citep{mmpred_lin_2019} trains an LSTM to predict the next event, time, and attributes in a many-to-many fashion, using what authors call a \textit{modulator}, which learns the importance of each attribute in the prediction of the event. \citep{attention_jalayer_2020} trains an encoder-decoder to predict the next activity based on two LSTMs coupled with an attention mechanism. Then, to predict the suffix, this model is used iteratively by always selecting the most probable activity. In ~\citep{adversarial_taymouri_2021} and~\citep{averages_ketyko_2022} ---the transformer variant--- the decoder is also trained to predict the same features that are used in the encoder, so it still accumulates errors from predicting features that are not relevant for the predictive task. While~\citep{mmpred_lin_2019} still has the aforementioned drawback, \citep{attention_jalayer_2020} avoids it by training an encoder-decoder to predict the next activity. However, the decoder only has information about the next activity during training and it is not trained specifically to predict the entire suffix. \citep{mann_khan_2021} adapts the Differentiable Neural Computer to predictive monitoring by separating the encoding and decoding phase of the neural model and preventing the decoder to write into the memory during the decoding phase. \citep{reinforcement_agarwal_2022} trains the GAN-LSTM proposed by \citep{next_taymouri_2020} for the next activity prediction problem by using reinforcement learning so that the predictions are compliant with a process model and take into account an estimated value of a KPI for that prediction.

In order to solve the drawbacks of the current state-of-the-art approaches, this paper proposes an encoder-decoder architecture with an additive attention mechanism to predict the activity suffix. The encoder receives information about the activities, times, and resources of the input prefix, whereas the decoder only predicts the activities of the suffix. The attention mechanism helps to glue these two components together by calculating the relevance of the learned context of the prefix w.r.t. each prediction from the decoder. During training, the decoder is trained using teacher forcing in order to speed up convergence, so it receives information from the whole suffix to predict while; whereas during inference a normalized beam search is used to minimize the exposure bias resulting from training with teacher forcing.

%%%%%%%%%%%%%%%%%%%%%%%%%
\section{Preliminaries}
\label{sec:preliminaries}
%%%%%%%%%%%%%%%%%%%%%%%%%

This section presents some basic concepts needed to understand our approach and the basic building blocks that will be used to construct our neural architecture to predict the activity suffix of a running case.

\begin{definition}[event]
    Let $A$ be the domain of activities, $C$ the domain of case identifiers, $T$ the domain of timestamps, and $R$ the domain of resources. An event, $e \in E$, is a tuple $(a, c, t, r)$, where $a \in A$, $c \in C$, $t \in T$, and $r \in R$.
\end{definition}

\begin{definition}[projection functions]
    The projection functions of an event in its activity, case identifier, and timestamp are denoted, respectively, by $\pi_A$, $\pi_C$, $\pi_T$, and $\pi_R$, i.e., $\pi_A(e) = a$, $\pi_C(e) = c$, $\pi_T(e) = t$, and $\pi_R(e) = r$.
\end{definition}

An ordered sequence of events that belongs to the same case is denoted as a trace.

\begin{definition}[trace]
    Let $S$ be the universe of traces, a trace $\sigma \in S$ is a non-empty sequence of events $\sigma = \langle e_1, \ldots, e_n \rangle$, which verifies that $\forall e_i, e_j \in \sigma; i, j \in \left[1, n\right]: (j > i) \land (\pi_C(e_i) = \pi_C(e_j)) \land (\pi_{T}(e_j) \geq \pi_{T}(e_i))$ where $|\sigma| = n$.
\end{definition}

Given the previous definition of trace, if two events have exactly the same timestamp, they are ordered following the already found order in the event log.

\begin{definition}[event log]
    An event log is a subset of the universe of traces, i.e., $L = \{\sigma_1, \ldots, \sigma_l \} \subseteq S$, where $|L| = l$, is the number of traces in the event log.
\end{definition}

To illustrate these concepts, let~\tablename~\ref{tab:log} serve as an example of an event log that records the steps performed in the process of patient management in a hospital. In this event log, there are five different activities, each one representing a step in the process. For each event ---row in the table---, the case identifier, activity, timestamp, and the resource assigned to its execution, are recorded. The sequence of events with the same case identifier is called a trace. In this example, there are two traces, with case identifiers \textit{Case1934} and \textit{Case2032}. The events for each trace appear chronologically ordered according to their timestamp. Note that the only mandatory elements in an event log are the case identifiers, the activities, and a partial order over them within each trace.

\begin{table}[t!]
    \centering
    \caption{Potential example of an event log from a hospital.}
    \begin{tabular}{lWlSlWl}
    \hline
    Case ID & Activity & Timestamp & Resource \\ \hline
    Case1934 & Registration & 15-02-2021 08:34:20 & Max \\ 
    Case1934 & CT scan & 16-02-2021 12:35:12 & Lewis \\
    Case1934 & Consultation & 20-03-2021 11:23:53 & Nicholas \\ 
    Case1934 & Blood analysis & 21-03-2021 10:54:12 & Sergio \\ \hline
    Case2032 & Registration & 17-02-2021 12:57:25 & Charles \\
    Case2032 & Blood analysis & 17-02-2021 13:05:24 & Carlos \\
    Case2032 & Discharge & 17-02-2021 18:46:26 & Fernando \\ \hline
    \end{tabular}
    \label{tab:log}
\end{table}

Predictive monitoring approaches work by using prefixes of events as their inputs.

\begin{definition}[event prefix and suffix]
    Let $\sigma \in S$ be a trace from the universe of traces $S$ such as $\sigma = \langle e_1, \ldots, e_n \rangle$ and $k \in \{1, \ldots, n\}$. Then, the k-event prefix is defined as $hd^k(\sigma) := \langle e_1, \ldots, e_k \rangle$ and the k-event suffix is defined as $tl^k(\sigma) := \langle e_{k+1}, \ldots, e_n \rangle$.
\end{definition}

The activity, timestamp, and resource of both the event prefix and suffix can be obtained by applying the projection functions $\pi_A$, $\pi_T$, and $\pi_R$ to $hd^k(\sigma)$ and $tl^k(\sigma)$, respectively. Thus, let us first define the next activity prediction problem, which can be dealt with as the prediction of the first suffix activity, $hd^{1}(td^{k}(\sigma)$:

\begin{definition}[next activity prediction problem]
    Let $hd^k(\sigma) = \langle e_1, \ldots, e_k \rangle$ be a prefix of length $k$. The aim of the next activity prediction problem is to find a function $\Omega_A$ such that $\Omega_A(hd^k(\sigma)) = \pi_{A}(hd^{1}(td^{k}(\sigma)))$, where $\pi_A$ is a projection function for the activities.
\end{definition}

Accordingly, the suffix prediction problem can be defined as a generalization of the next activity prediction problem, where the objective is to predict the $k$ activities of the whole suffix, $hd^{k}(td^{k}(\sigma)$, not only the first suffix activity:

\begin{definition}[activity suffix prediction problem]
    Let $hd^k(\sigma)$ be a prefix of length $k$. The aim of the suffix prediction problem is to find a function $\Omega_{SA}$ such as $\Omega_{SA}(hd^k(\sigma)) = \pi_{A}(td^k(\sigma))$.
\end{definition}

However, the general problem of the activity suffix prediction problem can be approached by iteratively predicting the next activity until the end of the trace is found. Formally, this type of prediction can be defined as follows:

\begin{definition}[iterative activity suffix prediction]
    The function $\Omega_{SA} \;$ can be constructed by iteratively predicting the next activity using $\Omega_A$ following the recursive definition: $(1) \; \Omega_{SA}(hd^k(\sigma)) = \sigma$ if $\Omega_{A}(hd^k(\sigma)) = [EOC]$, $(2) \; \Omega_{SA} = \Omega_{SA}(hd^k(\sigma) \oplus e)$, with $\pi_{A}(e) = \Omega_A(hd^k(\sigma))$, otherwise.
\end{definition}

This approach for predicting the suffix accumulates errors in each prediction of the next activity. Therefore, an alternative approach that may avoid this problem could be to learn a function $\Omega_{SA}$ that, given a certain prefix, directly returns the whole predicted suffix thus avoiding relying on a previously learned $\Omega_A$.

The main difference between both approaches is that for the direct prediction, the function $\Omega_{SA}$ is trained with the information of the whole suffix ---either implicitly, if predicting the whole suffix as a category, or explicitly, if predicting the suffix as a sequence of activities---, in contrast with the iterative approach, in which $\Omega_{SA}$ is trained only to predict the next activity. The approach presented in this paper falls in the direct prediction category due to it being trained with the whole suffixes at once and not only the next activities.

%%%%%%%%%%%%%%%%%%%%%%%%%%%%%%%%%%%%%%%%%%%%%%%%
\section{Deep learning architecture description}
\label{sec:abasp}
%%%%%%%%%%%%%%%%%%%%%%%%%%%%%%%%%%%%%%%%%%%%%%%%

%%%%%%%%%%%%%%%%%%%%%%%%%%%%%%%%%%%%%%%%%%%%
\subsection{ASTON architecture for training}
%%%%%%%%%%%%%%%%%%%%%%%%%%%%%%%%%%%%%%%%%%%%

\begin{figure}[ht!]
    \centering
    \includegraphics[width=0.8\textwidth]{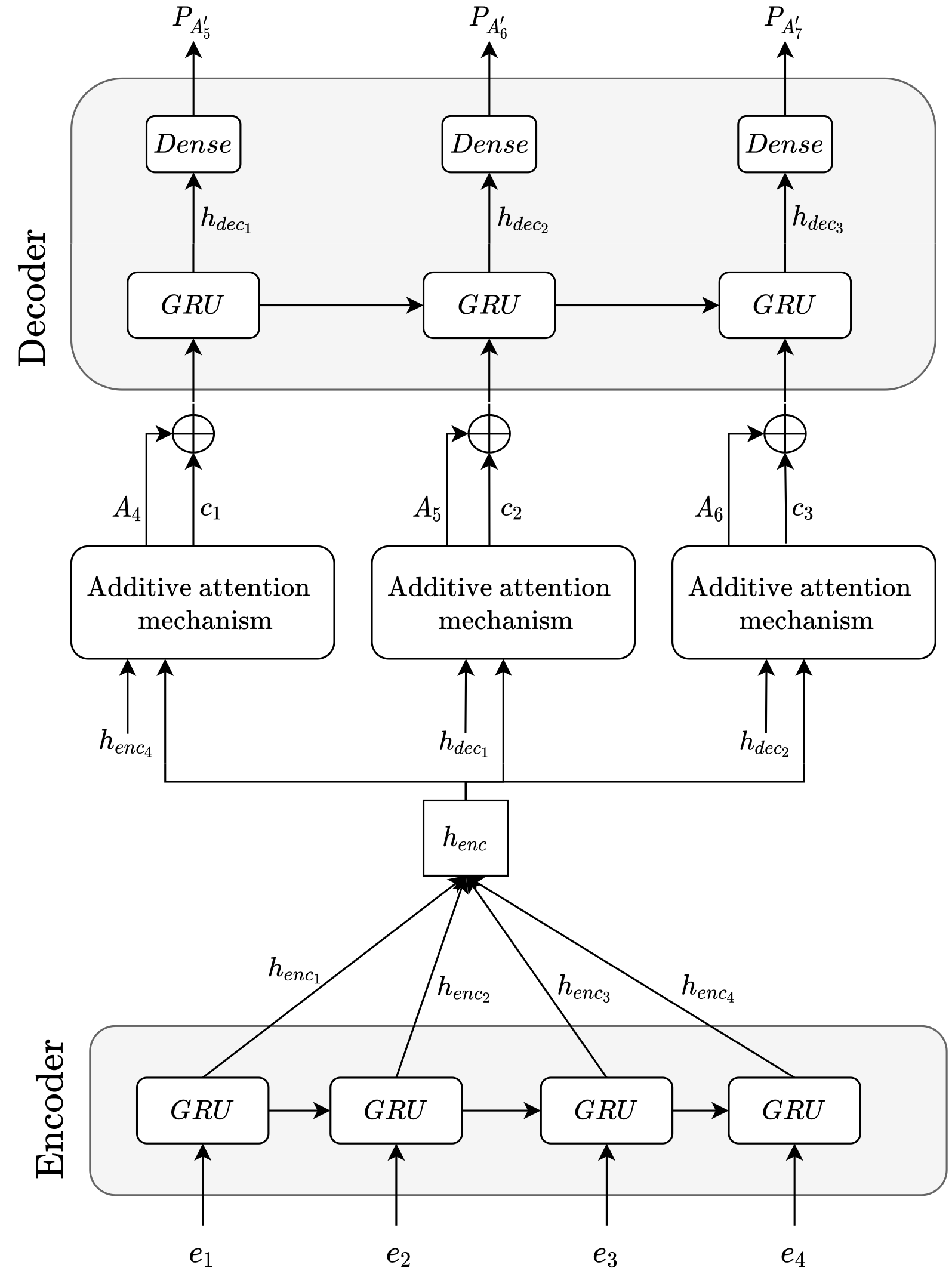}
    \caption{ASTON architecture for the training phase.}
    \label{fig:architecture}
\end{figure}

This section describes our approach, ASTON (\textbf{A}ctivity \textbf{S}uffix predic\textbf{T}i\textbf{O}n based on e\textbf{N}coder-decoder), to predict the activity suffix from a given input prefix using an encoder-recurrent neural model with an attention mechanism. \figurename~\ref{fig:architecture} shows how the ASTON neural model works during the training phase. ASTON architecture is built upon three main modules: \textit{(i)} the encoder, which learns an effective representation of the input prefix; \textit{(ii)} the decoder, which predicts the suffix using the information from the encoder; and \textit{(iii)} the attention mechanism, which helps glue together the encoder and the decoder by computing how relevant is a given state from the decoder w.r.t. the input prefix in order to make a prediction.

The encoder is implemented as a Gated Recurrent Unit (GRU)~\citep{Cho2014} that learns from the input prefix, which contains information about not only the activities ---$A_i$ in \figurename~\ref{fig:architecture}--- but also the time and resources for each event of the prefix, $e_i$. The output of this module is a sequence of hidden states for each event of the prefix, $h_{enc}$, that will be used by the decoder and the attention mechanism to make the suffix prediction.

The decoder is also implemented as a GRU and outputs a sequence of hidden states, $h_{dec_n}$, which, in turn, are projected into the probability distribution of the next activity of the suffix, $P'{A_n}$. The input in each timestep is the concatenation of the context vector computed using the attention mechanism and the previous activity. In order to speed up the convergence and training of the predictive model, \textit{teacher forcing}~\citep{goodfellow_deep_2016} is employed, i.e., the ground-truth activity $A_i$ in each timestep is fed instead of the predictions $A'_i$ of the decoder. Thus, the convergence of the neural network is greatly improved since, otherwise, the hidden states would be updated with potentially wrong predictions. Note that to predict the first event of the prefix, the decoder is fed with the last event of the prefix ---$A_4$ in \figurename~\ref{fig:architecture}---.

The attention mechanism ties together the previous two components by computing a context vector $c_i$ that represents the relevance of each hidden state of the decoder, $h_{dec_n}$, w.r.t. the full sequence of hidden states of the encoder, $h_{enc_n}$. Note that for the first event of the decoder, no previous hidden states of the decoder are available, so the last state of the encoder is used ---$h_{enc_4}$ in \figurename~\ref{fig:architecture}---. In this paper, an additive attention mechanism~\citep{bahdanau_neural_2016} is used, which will be detailed in subsequent sections.

%%%%%%%%%%%%%%%%%%%%%%%%%%%%%%
\subsubsection{Input encoding}
%%%%%%%%%%%%%%%%%%%%%%%%%%%%%%

ASTON encodes each event as a tensor resulting from the concatenation of the activity, with the time features computed from the timestamps, and the resources assigned to each activity. Activities are first encoded into integer identifiers and then into embeddings, represented by a vector assigned to each activity. Embeddings are used because they are less memory intensive than a one-hot encoding representation in event logs with a high number of activities or resources. Besides, since they are trained along the rest of the parameters of the neural network, they are susceptible to capturing the execution context of its associated activity.

On the other hand, the timestamps of the event cannot be used directly as inputs to the neural network, so the following time features are computed from these timestamps: the time elapsed between one event and its predecessor; the time elapsed between one event and the beginning of the trace; the time elapsed between one event and the midnight of the same day; and the month, the day of the week, and the hour of the day in which the event has happened. In addition, the time elapsed between one event and its predecessor and the time elapsed between one event and the start of the trace are standardized using the z-score technique~\citep{Singh2020}:

\begin{equation}
    Z = \frac{X - \hat{X}}{S}
\end{equation}

Where $X$ is a numerical variable, $\hat{X}$ is the sample mean of the variable, and $S$ is its sample standard deviation. However, the z-score standardization is applied after a logarithmic transformation since there exists a high degree of variability in these variables; so applying the logarithm amplifies the differences in the values near zero, which in turn helps to capture the dependencies among the activities, as shown in~\citep{camargo_learning_2019}. The other time-related variables are normalized using the maximum value for each type of variable. For example, the time since midnight is normalized by dividing it by 86 400, i.e., the number of seconds in a day.

\begin{figure}[t]
    \centering
    \includegraphics[width=1\textwidth]{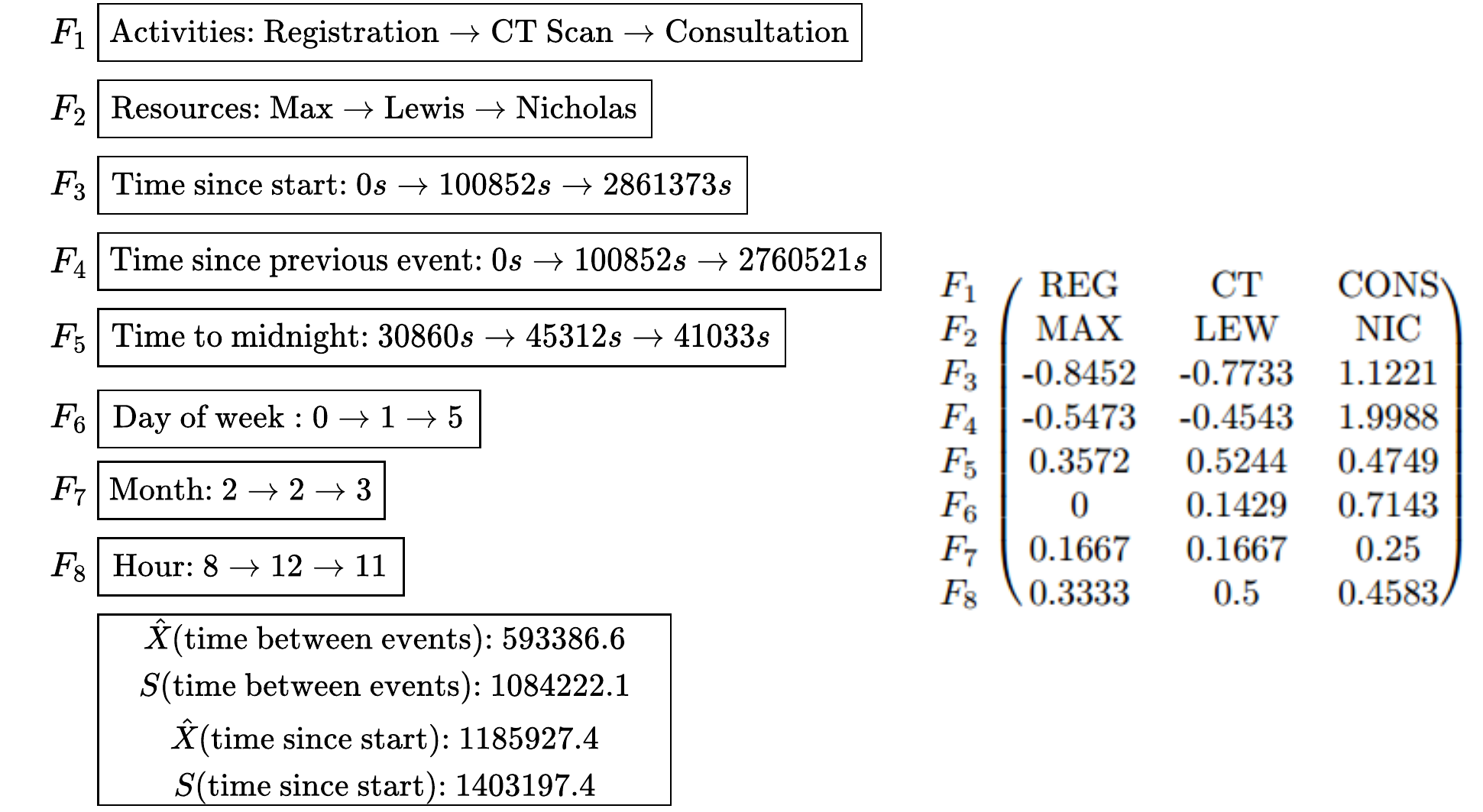}
    \caption{Encoding of the prefix corresponding to the three first events of the trace \textit{Case1934} from the log shown in \tablename~\ref{tab:log}. For the sake of clarity, categorical features are shown before the embedding phase.}
    \label{fig:encoding}
\end{figure}

\figurename~\ref{fig:encoding} shows an example of the tensor that encodes the information of a prefix. In this tensor, columns represent the prefix events, while rows represent the features computed for each prefix event. Thus, for a given column, i.e., for an event, the feature $F_1$ refers to the activity; $F_2$ indicates, if available, the resource; and $F_3$, $F_4$, $F_5$, $F_6$, $F_7$, and $F_8$ are the normalized time-related features calculated from the event timestamp, namely: the time since the beginning of the trace, the time since the previous event, the time since midnight, the day of the week ---starting from Monday as 0---, month, and hour, respectively.

%%%%%%%%%%%%%%%%%%%%%%%%%%%%%%%%%%%
\subsubsection{Attention mechanism}
\label{subsec:attention}
%%%%%%%%%%%%%%%%%%%%%%%%%%%%%%%%%%%

As explained in previous sections, ASTON uses an attention mechanism to tie together the encoder and the decoder modules of the architecture. The motivation behind this is that using only the last hidden state of the encoder might not be enough to obtain an accurate prediction because we would be assuming that this vector perfectly synthesizes the execution context of the prefix. Therefore, an attention mechanism is introduced to help leverage the complete information from the encoder.

Generally, an attention mechanism performs three steps in order to compute a \textit{context vector}, $c_i$, that summarizes the whole information of the encoder w.r.t. the current state of the decoder. These steps are the following:

\begin{itemize}
    \item First, an attention score function calculates the relevance of a hidden state of the decoder, $s_{i-1}$, w.r.t. the sequence of hidden states of the decoder, $h_t$. In ASTON, this score function is defined as the addition of the hidden states of the encoder and the decoder~\citep{bahdanau_neural_2016}. Therefore, the scoring operation is defined as follows:

    \begin{equation}
        e_{t, i} = w^T tanh(W_1 s_{t-1} + W_2 h_i)
    \end{equation}

    where $W_1$, $W_2$, and $w^T$ are learnable parameters; $s_{t-1}$ is the previous hidden state of the encoder; $h_i$ is the $i_{th}$ hidden state of the encoder; and $e_{t, i}$ is the calculated score. 
    
    \item The second step explicitly calculates which parts of the input data the model will attend to~\citep{attention_brauwers_2021}. Typically, the softmax function is applied to the scores, $e_{t}$ to compute the scores $\alpha_{t}$ in the following way:

    \begin{equation}
        \alpha_{t,i} = softmax(e_{t,i}) = \frac{exp(e_{t,i})}{\sum^E_{j=1} exp(e_{t,j})}
    \end{equation}
    
    where $E$ is the sequence of hidden encoder states. 
    
    \item Finally, a context vector, $c_t$, is calculated by combining the attention scores $\alpha$ with the hidden states of the encoder as follows:

    \begin{equation}
        c_t = \sum^E_{i=1} \alpha_{t, i} h_i
    \end{equation}
\end{itemize}

Once the context vector $c_t$ is calculated, the decoder input $I_t$ at each timestep $t$ is computed as follows:

\begin{equation}
    I_t = c_t \oplus A'_{t-1}
\end{equation}

where $c_t$ is the context vector calculated using the attention mechanism, $\oplus$ is the concatenation operator, and $A'_{t-1}$ is the predicted activity for the previous timestep ---or the last activity of the prefix if $t=0$---.

\begin{figure}[!t]
    \centering
    \includegraphics[width=0.79\textwidth]{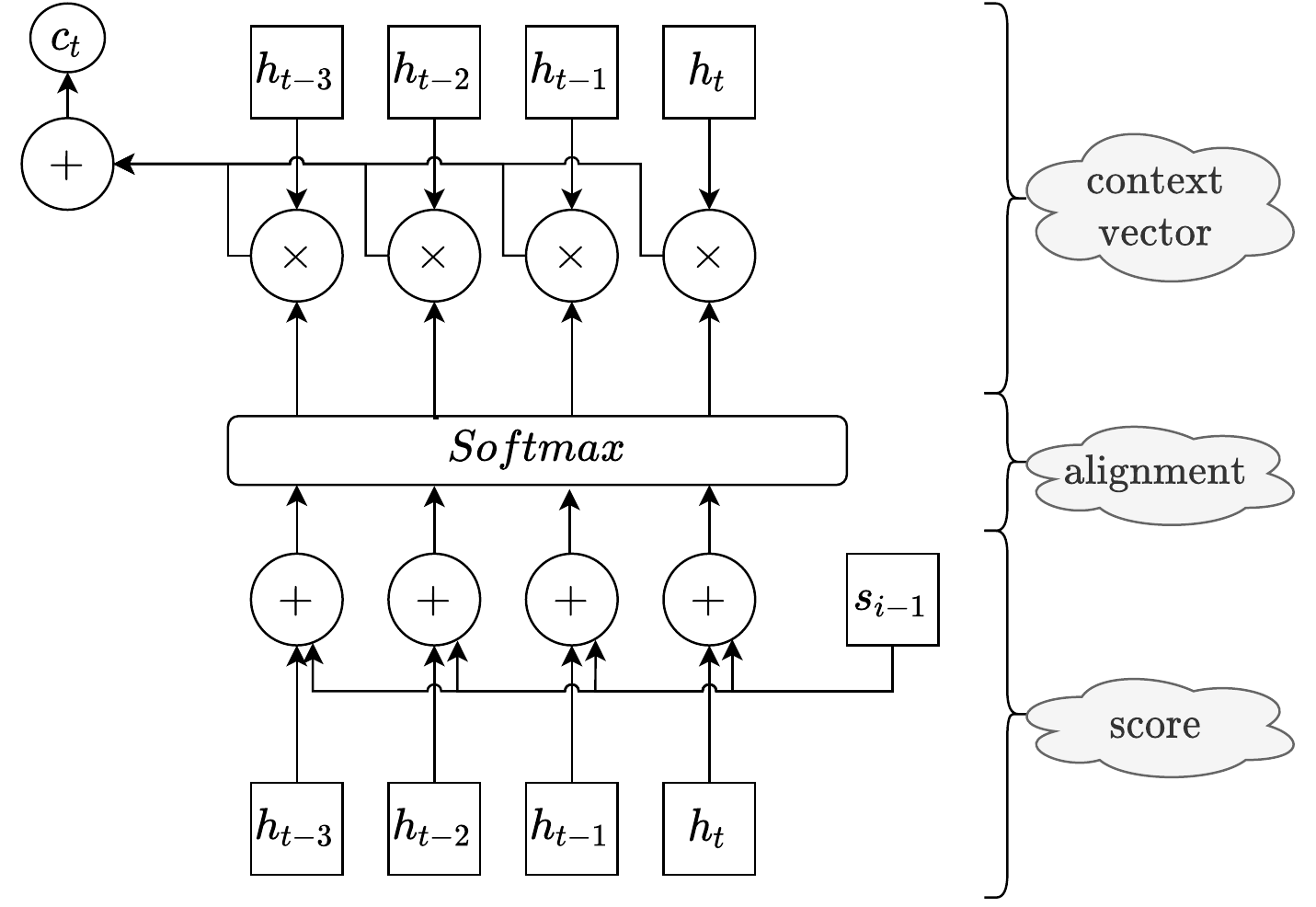}
    \caption{Attention mechanism of ASTON when processing an input prefix. The squares represent hidden states from the encoder, while $s_{t-1}$ is the last hidden state of the decoder, the circle with a $+$ inside represents the addition operation, and the circle with a $\times$ inside represents the product between two vectors.} 
    \label{fig:attention}
\end{figure}

Let \figurename~\ref{fig:attention} serves as a graphical illustration of the attention mechanism used in ASTON. Here, $h_{t-3}, ..., h_t$ are the hidden states of the encoder, $s_{t-1}$ is the last state of the decoder, and $c_t$ is the resulting context vector. The first stage of the attention mechanism, depicted in the bottom part of the picture, shows how the scores are calculated from the hidden states of the encoder and the last state of the decoder by adding them pairwise. The middle part shows how the alignment of the previously calculated scores is performed by applying a softmax operation to the previously calculated scores and, finally, the upper part shows how each of the hidden states of the encoder is combined with the aligned scores by multiplying and adding them to produce the final context vector, $c_t$.

%%%%%%%%%%%%%%%%%%%%%%%%%%
\subsubsection{Training}
\label{subsubsec:training}
%%%%%%%%%%%%%%%%%%%%%%%%%%

As the activity suffix prediction can be viewed as a multiclass prediction problem of a sequence of activities, ASTON is trained using the cross-entropy loss function defined for activities. Thus, ASTON applies the cross-entropy loss to each predicted activity of the suffix. Formally, let $a$ be a ground-truth activity index, $a'$ its corresponding predicted probability distribution, and $C$ be the number of different activities in the event log. Then, the cross-entropy loss for activities is defined as follows:

\begin{equation}
l(a, a') = - \log \frac{exp(a'_{a})}{\sum^C_{c=1} exp(a'_c)}
\end{equation}

Then, the aforementioned loss function is extended to the whole predicted suffix as follows:

\begin{equation}
L(A, A') = \sum^{K}_{k=1} l(A_k, A'_k)
\end{equation}

Where $A$ is the sequence of ground-truth activity indices of the suffix, $A'$ is the sequence of probability distributions of each predicted activity, and $K$ is the length of the suffix. Note that both $A$ and $A'$ are padded with end-of-case tokens to the maximum length of the log trace, so they are guaranteed to have the same length during training. However, these pad values are not masked in the loss function, since in the experimentation predictions were more stable in this way.

%%%%%%%%%%%%%%%%%%%%%%%%%%%%%%%%%%%%%%%%%%%%%
\subsection{ASTON architecture for inference}
\label{subsec:inference}
%%%%%%%%%%%%%%%%%%%%%%%%%%%%%%%%%%%%%%%%%%%%%

\begin{figure}[ht!]
    \centering
    \includegraphics[width=0.8\textwidth]{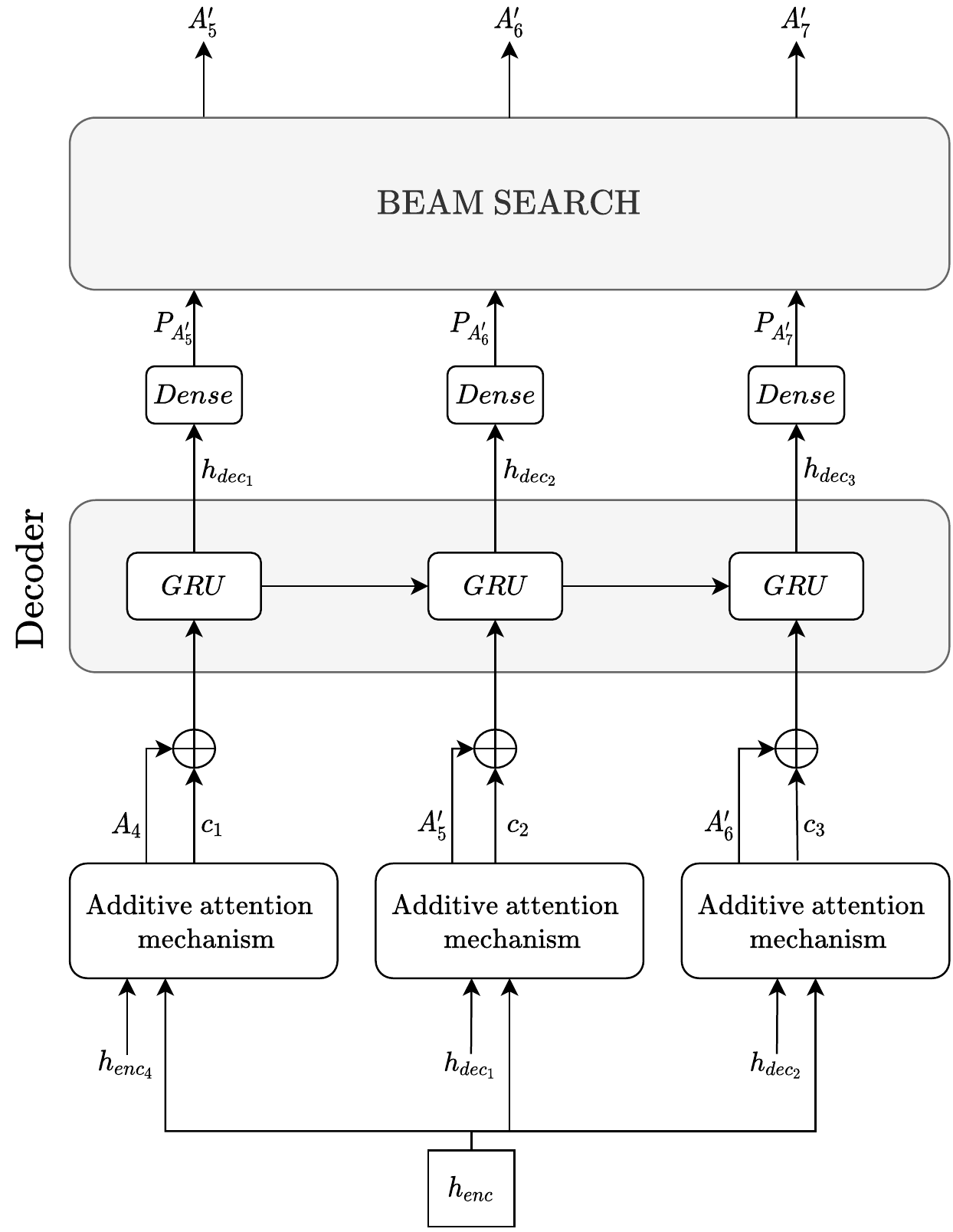}
    \caption{ASTON architecture for the inference phase, where the encoder is omitted for clarity.}\label{fig:architecture-2}
\end{figure}

During the inference phase, the inputs of ASTON differ slightly, as \figurename~\ref{fig:architecture-2} shows. This is due to the fact that the decoder cannot use the ground truth activities as inputs ---they are only available at the training phase---, meaning that they must be substituted by the predicted activities, $A'_t$, except for the first activity of the decoder, which is the last activity of the input prefix ---known during inference---. Thus, applying teacher forcing causes an \textit{exposure bias}, since during training ground truth inputs are used, whereas during inference the predicted outputs are used as inputs in subsequent timesteps. Therefore, how the next activity of the suffix is determined during inference is an issue of capital importance.

At each point of the sequence, the output of the decoder is not an activity but a probability distribution over the set of activities $P'_{A'_{t}}$, and the predicted activity $A'_t$ must be selected from this probability distribution. To illustrate how this selection could be performed, \figurename~\ref{fig:decoder-probas} shows the output probabilities generated from the decoder after predicting the following two activities of a given prefix. Thus, in predicting process monitoring, the most commonly used strategies for the activities selection are the following:

\begin{figure}[!b]
    \centering
    \includegraphics[width=0.8\textwidth]{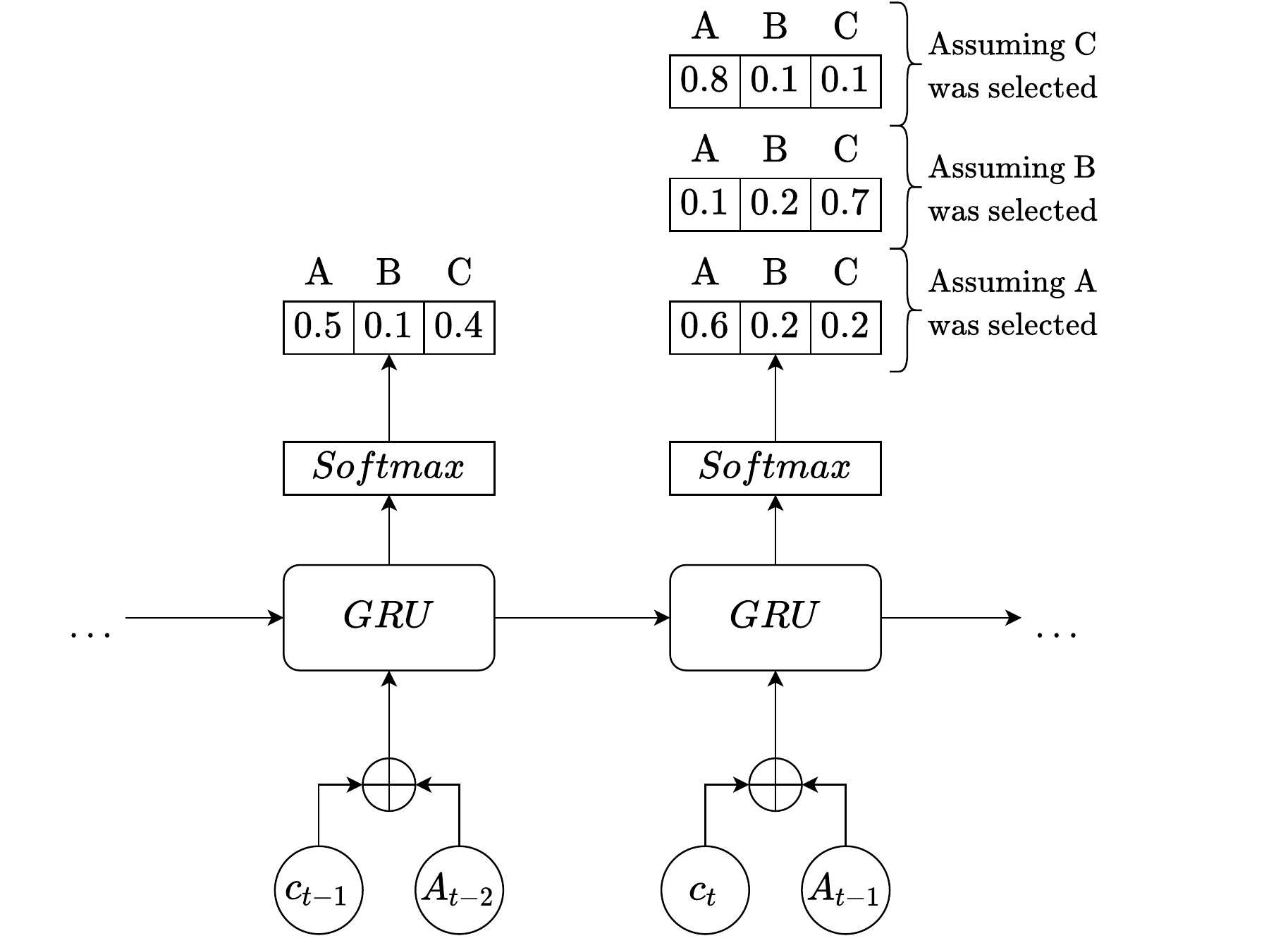}
    \caption{Output probabilities for the next two activities of a given prefix, where $c_t$ represents the context at timestep $t$ computed with the attention mechanism, $A_t$ represents the embedding for a predicted activity at timestep $t$, $\oplus$ is the concatenation operator, and the numbers in the top of the figure represent the probability of a given activity at a certain timestep. Depending on which activity is selected on timestep $A_{t-2}$ ---A, B, or C---, three different probability distributions for timestep $A_{t-1}$ can be selected.}
    \label{fig:decoder-probas}
\end{figure}

\begin{itemize}
    \item \textit{Argmax}~\citep{camargo_learning_2019, Tax2017}, where the most likely activity is always selected at each step of the prediction. In the example of \figurename~\ref{fig:decoder-probas}, the activities ``A'' and ``A'' would be selected, getting the suffix ``AA'', with a joint probability of $0.5 \cdot 0.6 = 0.3$. As noted in~\citep{rama-maneiro_deep_2022}, this strategy works reliably in event logs with short traces while underperforming in logs with long ones. Note that, since the training is performed using the full suffix, the predictions can also be obtained in a one-shot way ---with a single call to the model---, but the suffixes selected would always correspond with this strategy because the predictive model is trained by maximum likelihood estimation.
     
    \item \textit{Random sampling}~\citep{camargo_learning_2019, Evermann2017}, where the next activity is randomly sampled from the probability distribution of the next activities. Thus, recalling the example of \figurename~\ref{fig:decoder-probas}, we might first select ``B'' as the next activity and then ``A'' as the final prediction. The final predicted suffix, ``BA'', would have a joint conditional probability of $0.1 \cdot 0.8 = 0.08$. As noted in~\citep{rama-maneiro_deep_2022}, this strategy, unlike \textit{argmax}, works better in logs with long traces while underperforming in logs with short ones. In this case, the suffix prediction cannot be obtained in a one-shot fashion because in each timestep the activity is randomly sampling, which affects the subsequent predictions.
\end{itemize}

\begin{figure}[!b]
    \centering
    \includegraphics[width=0.8\textwidth]{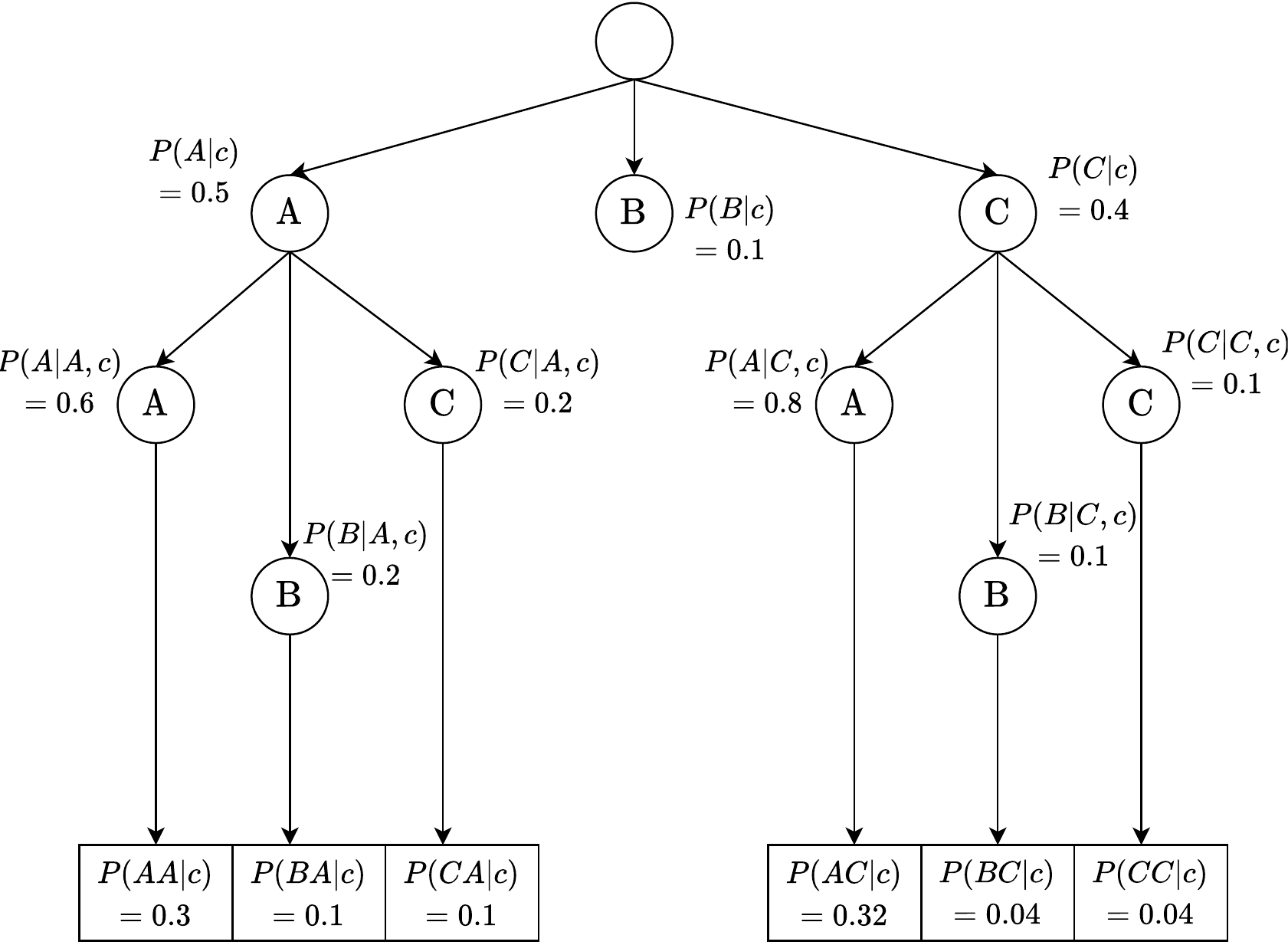}
    \caption{Probability tree for a suffix of two activities and three possible predictions at each step.}
    \label{fig:beam-search}
\end{figure}

As shown in the aforementioned examples, the discussed strategies might end up not selecting the suffix with the highest probability overall ---which is ``AC''---. On the one hand, the \textit{argmax} strategy does not explore the solution space properly, being prone to selecting a suboptimal solution and/or getting stuck repeating the same set of activities without ever reaching the end of the case token. On the other hand, the \textit{random sampling} strategy allows too much freedom during inference, which tends to predict suboptimal suffixes. These shortcomings are due to the fact that both strategies select only one activity at each step of the prediction and, therefore, the predicted suffix might not be the one with the highest joint probability.

To deal with the drawbacks of these two strategies, as \figurename~\ref{fig:architecture-2} shows, ASTON uses the beam search~\citep{bengio_scheduled_2015}, which is an optimization of the exhaustive search that reduces the memory and time needed to obtain a solution. Particularly, the complexity of the beam search is $\mathcal{O}{(B_w \cdot |A| \cdot |\sigma|)}$, where $B_w$ is the beam width, indicating the degree of exploration of the algorithm; $|A|$ is the number of different activities in the log; and $|\sigma|$ is the maximum length of a suffix. As an example of the beam search algorithm,~\figurename~\ref{fig:beam-search} presents the graph obtained from the output probabilities shown in ~\figurename~\ref{fig:decoder-probas}. For $B_w = 2$, only the two most probable activities are expanded at each step. Thus, in the first step ``A'' and ``C'' activities are selected since they have the two highest probabilities, 0.5 and 0.4, respectively; whereas ``B'' is not considered to explore possible suffixes because its probability is 0.1. Note that the expansions are performed in order of probability, i.e., ``A'' is expanded before ``C'' due to it is more probable than ``C''. In the second step, no more expansions are made since the end of the suffix prediction is reached, and, as the search stops, the suffix with the highest joint-conditioned probability is selected, which, in this example, is ``AC''.

Note that beam search, unlike both the argmax and random sampling strategies, selects the suffix with the highest joint probability instead of getting stuck in a suboptimal solution. In particular, the joint conditioned probability of a partial suffix is formally computed through the following expression:

\begin{equation}
    log P(y_i, \ldots, y_0| c) = \sum^{i}_{j = 0} log P(y_j | y_{j-1}, \ldots, y_0, c)
    \label{eq:beam-2} 
\end{equation}

Where $c$ is the prefix that conditions the prediction; $y_i$ is the $i_{th}$ activity, with $i \in \{0, \ldots, N - 1\}$; and the logarithm operation is applied to deal with the numerical instability of consecutive products of floating-point numbers in the interval $[0, 1]$.

Nevertheless, the expression~\ref{eq:beam-2} tends to favor short suffixes since a negative log probability is added for each predicted event, yielding a lower score for the predicted suffix~\citep{wu_googlenmt_2016}. Therefore, we have adopted the same normalization strategy as in~\citep{wu_googlenmt_2016} to fairly compare suffixes of different lengths, resulting in the following expression:

\begin{equation}
    log P(y_i, \ldots, y_0| c) = \frac{(5 + 1)^\alpha}{(5 + i)^\alpha} \sum^{i}_{j = 0} log P(y_j | y_{j-1}, \ldots, y_0, c)
    \label{eq:beam-3}
\end{equation}

Where $i$ is the current prediction step and $\alpha$ is a factor which is set to $0.65$ in our experimentation. This suffix normalization is what distinguishes the beam search implemented in ASTON from the beam search proposed by~\citep{Francescomarino2017} and~\citep{adversarial_taymouri_2021}. Thus, \citep{Francescomarino2017} does not perform any kind of normalization to the output probabilities, whereas~\citep{adversarial_taymouri_2021}, instead of normalizing, stops the prediction if the suffix length exceeds a certain threshold related to the average trace length of the log. Therefore, their stopping strategy is too restrictive by not allowing the prediction of suffixes longer than this threshold.

%%%%%%%%%%%%%%%%%%%%%%
\section{Evaluation}
\label{sec:evaluation}
%%%%%%%%%%%%%%%%%%%%%%

%%%%%%%%%%%%%%%%%%%%%%%%%%%%%%%
\subsection{Experimental setup}
%%%%%%%%%%%%%%%%%%%%%%%%%%%%%%%

We compared ASTON against 5 other state-of-the-art approaches~\citep{adversarial_taymouri_2021, Tax2017, Evermann2017, Francescomarino2017, camargo_learning_2019} using the same experimental conditions as in~\citep{rama-maneiro_deep_2022}. From these approaches, \citep{camargo_learning_2019} supports two different modes of operation, so two results were reported in this case. ASTON has been configured with the same hyperparameters for every event log since we found during the experimentation that the ASTON architecture is pretty robust to hyperparameter changes. Thus, the neural network has been trained during 150 epochs, using a batch size of 64; 32 for the dimensionality of the embedding; a 2-layer GRU with a hidden dimension of 32; a dropout of 0.1 in both the encoder and the decoder; and the Adam optimizer with a learning rate of 0.005. Furthermore, in the inference phase, a beam width of 5 has been set for the predictions of the suffix due to its good empirical performance on the validation sets of the tested event logs.

The state-of-the-art approaches have been evaluated on the same event logs as considered in~\citep{rama-maneiro_deep_2022} but, to provide a fair comparison, no results were reported for the approach~\citep{camargo_learning_2019} in the event logs \textit{Nasa} and \textit{Sepsis} because they do not contain information about resources and, therefore, cannot be executed in those logs. \tablename~\ref{tab:statistics} shows some relevant statistics for the event logs used in the experimentation. It is remarkable to note the high variability among the trace length since it increases the uncertainty in predicting the end of the trace for a given prefix. Moreover, there is also a high variability in the number of trace variants, which results in more difficulties during the training phase of the neural network for identifying common patterns between executions. In addition, we performed the evaluation by following a 5-fold cross-validation, splitting each training fold in an 80\%--20\% fashion to obtain a training and validation set ---the validation set is used to select the best-performing model in terms of its validation loss---.

\begin{table*}[!b]
    \centering
    \scriptsize
    \caption{Statistics of the event logs used in the evaluation. Time-related measures are shown in days.}
    \begin{tabular}{l|cccScScScScScScSc}
        Event log & \rotatebox{90}{Traces} & \rotatebox{90}{Activities} & \rotatebox{90}{Events} & \rotatebox{90}{\shortstack[l]{Avg. case \\ length}} &
        \rotatebox{90}{\shortstack[l]{Max. case \\ length}} & \rotatebox{90}{\shortstack[l]{Avg. event \\ duration}} &
        \rotatebox{90}{\shortstack[l]{Max. event \\ duration}} & \rotatebox{90}{\shortstack[l]{Avg. case \\ duration}} &
        \rotatebox{90}{\shortstack[l]{Max. case \\ duration}} & \rotatebox{90}{Variants} \\ \hline
        BPI 2012 & 13087 & 36 & 262200 & 20.04 & 175 & 0.45 & 102.85 & 8.62 & 137.22 & 4366 \\
        BPI 2012 A & 13087 & 10 & 60849 & 4.65 & 8 & 2.21 & 89.55 & 8.08 & 91.46 & 17 \\
        BPI 2012 C. & 13087 & 23 & 164506 & 12.57 & 96 & 0.74 & 30.92 & 8.61 & 91.46 & 4336 \\
        BPI 2012 O & 5015 & 7 & 31244 & 6.23 & 30 & 3.28 & 69.93 & 17.18 & 89.55 & 168 \\
        BPI 2012 W & 9658 & 19 & 170107 & 17.61 & 156 & 0.7 & 102.85 & 11.69 & 137.22 & 2621 \\
        BPI 2012 W C. & 9658 & 6 & 72413 & 7.5 & 74 & 1.75 & 30.92 & 11.4 & 91.04 & 2263 \\
        BPI 2013 C. P. & 1487 & 7 & 6660 & 4.48 & 35 & 51.42 & 2254.84 & 178.88 & 2254.85 & 327 \\
        BPI 2013 I. & 7554 & 13 & 65533 & 8.68 & 123 & 1.57 & 722.25 & 12.08 & 771.35 & 2278 \\
        Env. permit & 1434 & 27 & 8577 & 5.98 & 25 & 1.09 & 268.97 & 5.41 & 275.84 & 116 \\ 
        Helpdesk & 4580 & 14 & 21348 & 4.66 & 15 & 11.16 & 59.92 & 40.86 & 59.99 & 226 \\
        Nasa & 2566 & 94 & 73638 & 28.7 & 50 & 0.00 & 0.00 & 0.00 & 0.00 & 2513 \\ 
        Sepsis & 1049 & 16 & 15214 & 14.48 & 185 & 2.11 & 417.26 & 28.48 & 422.32 & 845 \\ \hline
    \end{tabular}
    \label{tab:statistics}
\end{table*}

To compare the results of the state-of-the-art approaches, the same evaluation metric as in~\citep{rama-maneiro_deep_2022} has been used: the Damerau-Levenshtein normalized distance \citep{Gali2019}. The Damerau-Levenshtein distance, $lev_{a, a'}(|a|, |a'|)$, computes the edit distance between a pair of suffixes $a$ and $a'$, i.e., how many substitutions, transpositions, deletions, and insertions are needed to transform $a$ into $a'$. However, since suffixes are of variable lengths, it is more convenient to normalize this metric by the maximum suffix length, obtaining the Damerau-Levenshtein normalized distance as follows:

\begin{equation}
    DL\_norm(a, a') = 1 - \frac{lev_{a, a'}(|a|, |a'|)}{max(|a|, |a'|)}
\end{equation}

Finally, to quantify the differences among the state-of-the-art approaches and evaluate whether the results they obtain are statistically different, a two-stage statistical comparison has been applied. Following the procedure defined in ~\citep{rama-maneiro_deep_2022} and~\citep{rama-maneiro_embedding_2021},  Bayesian statistical tests instead of null hypothesis statistical tests were applied due to their easiness of interpretation and robustness in quantifying the differences among the approaches. In the first stage of the statistical test pipeline, a Bayesian approach to rank the approaches using the Plackett-Luce model~\citep{calvo_bayesian_2018} is applied, allowing the classification 
 of multiple approaches based on their probability of being the best predictor. Then, in the second stage of the statistical test pipeline, a Bayesian hierarchical test~\citep{benavoli_time_2017} between the three best approaches according to the previous ranking is performed. This second stage is especially insightful because, unlike the Plackett-Luce model, it uses the results obtained for each fold of the cross-validation evaluation procedure, so it considers that an approach can perform badly on one fold but very well on the rest, and vice versa. In order to perform these statistical tests, the library \texttt{scmamp}~\citep{calvo_scmamp_2016} is used.

%%%%%%%%%%%%%%%%%%%%
\subsection{Results}
%%%%%%%%%%%%%%%%%%%%

\tablename~\ref{tab:results} shows the results of the experimentation. ASTON obtains the best results in 5 out of 12 event logs, the second best result in 6 out of 12, and the third best result in 1 out of 12. Therefore, ASTON is mostly either the best or the second-best approach from the state of the art. These results seem to indicate that ASTON is highly consistent in both event logs with shorter and longer traces: specifically, for the five logs in which it obtains the best results, there exists a significant difference in performance when comparing with the second-best approach in each log ---ranging from 0.0113 until 0.0787, with an average of 0.0534---; whereas for the other seven logs, it obtains results that are very close to the best approach in each log ---with an average difference of 0.0242---.

\begin{table}[!b]
    \small
    \centering
    \caption{Results of the 5-fold cross-validation for the activity suffix prediction task in terms of the Damerau-Levenshtein similarity. The best, second-best, and third-best results are highlighted in green, orange, and yellow, respectively.}
    \begin{tabular}{l|cWcWcWcWcWcWcWc}
        Event log & \rotatebox{90}{\shortstack[l]{Camargo/ \\ argmax}} & \rotatebox{90}{\shortstack[l]{Camargo/ \\ random}} & \rotatebox{90}{Evermann} & \rotatebox{90}{\shortstack[l]{Francesco- \\marino}} & \rotatebox{90}{Tax} & \rotatebox{90}{Taymouri} & \rotatebox{90}{ASTON} \\ \hline
        BPI 2012 & 0.1851 & \cellcolor{orange}{\textbf{0.3891}} & \cellcolor{yellow}{\textbf{0.1986}} & 0.1321 & 0.1409 & 0.1867 & \cellcolor{ForestGreen}{\textbf{0.4004}} \\ 
        BPI 2012 A & \cellcolor{ForestGreen}{\textbf{0.6536}} & \cellcolor{orange}{\textbf{0.6441}} & 0.5847 & 0.2871 & 0.4597 & 0.5420 & \cellcolor{yellow}{\textbf{0.6340}} \\ 
        BPI 2012 C. & 0.2218 & \cellcolor{ForestGreen}{\textbf{0.4495}} & 0.2693 & 0.0883 & 0.1717 & \cellcolor{yellow}{\textbf{0.3557}} & \cellcolor{orange}{\textbf{0.4423}} \\ 
        BPI 2012 O & \cellcolor{ForestGreen}{\textbf{0.6845}} & \cellcolor{yellow}{\textbf{0.6042}} & 0.5544 & 0.4591 & 0.4972 & 0.5150 & \cellcolor{orange}{\textbf{0.6087}} \\ 
        BPI 2012 W & 0.1941 & \cellcolor{orange}{\textbf{0.3110}} & \cellcolor{yellow}{\textbf{0.2800}} & 0.1410 & 0.0975 & 0.2493 & \cellcolor{ForestGreen}{\textbf{0.3897}} \\ 
        BPI 2012 W C. & 0.0501 & 0.3191 & \cellcolor{ForestGreen}{\textbf{0.3372}} & 0.1126 & 0.0789 & \cellcolor{yellow}{\textbf{0.3201}} & \cellcolor{orange}{\textbf{0.3366}} \\ 
        BPI 2013 C. P. & \cellcolor{orange}{\textbf{0.6641}} & 0.5337 & \cellcolor{yellow}{\textbf{0.6416}} & 0.5276 & 0.5824 & 0.4358 & \cellcolor{ForestGreen}{\textbf{0.7042}} \\ 
        BPI 2013 I. & 0.2607 & \cellcolor{orange}{\textbf{0.5294}} & \cellcolor{yellow}{\textbf{0.4730}} & 0.3607 & 0.3336 & 0.2452 & \cellcolor{ForestGreen}{\textbf{0.6069}} \\ 
        Env. permit & \cellcolor{ForestGreen}{\textbf{0.8440}} & 0.7595 & 0.5713 & 0.3924 & \cellcolor{yellow}{\textbf{0.8163}} & 0.4271 & \cellcolor{orange}{\textbf{0.8268}} \\ 
        Helpdesk & \cellcolor{ForestGreen}{\textbf{0.9110}} & 0.8524 & 0.8354 & 0.4619 & \cellcolor{yellow}{\textbf{0.8695}} & 0.6298 & \cellcolor{orange}{\textbf{0.8919}} \\ 
        Nasa & - & - & 0.1218 & 0.0842 & \cellcolor{yellow}{\textbf{0.2320}} & \cellcolor{ForestGreen}{\textbf{0.4844}} & \cellcolor{orange}{\textbf{0.4545}} \\ 
        Sepsis & - & - & \cellcolor{yellow}{\textbf{0.2693}} & 0.0742 & 0.1158 & \cellcolor{orange}{\textbf{0.3217}} & \cellcolor{ForestGreen}{\textbf{0.3811}} \\ \hline
    \end{tabular}
    \label{tab:results}
\end{table}

\begin{figure*}[!b]
    \centering
    \includegraphics[width=1\textwidth]{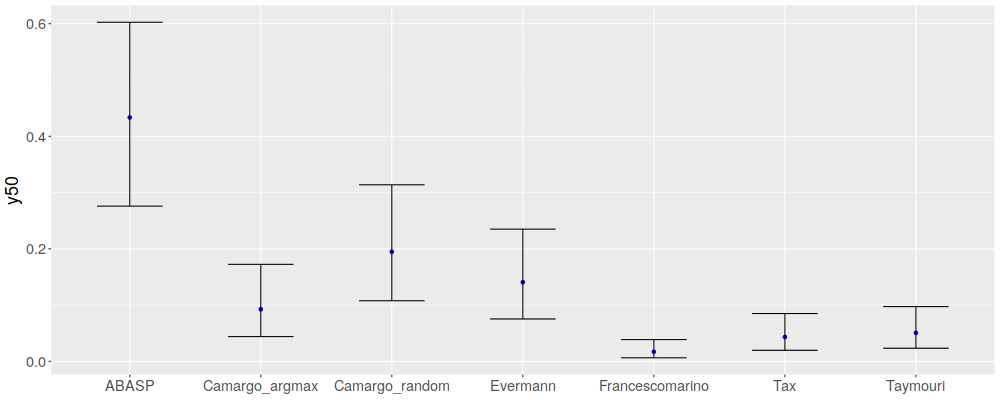}
    \caption{Credible intervals and the expected probability of winning for the tested approaches.}
    \label{fig:credible}
\end{figure*}

However, it is difficult to evaluate which is the best approach overall, since it does not only depend on the number of event logs in which each approach is the best but also on the quantitative differences between the results obtained in each event log. Thus, to rank the approaches based on their performance, a Bayesian Plackett-Luce model is applied, the results of which are shown in \tablename~\ref{tab:plackett-luce}. ASTON obtains the best overall ranking and the highest probability of being the best approach ---43.55\%---, with a difference of 23.5 percentage points over the second approach ---Camargo/random---. Furthermore, \figurename~\ref{fig:credible} shows the credible intervals ---quantiles 5\% and 95\%--- and the expected probability of being the best approach. In this figure, a non-overlap of the credible intervals of a pair of approaches means that they are statistically different for the given significance level. As shown in \figurename~\ref{fig:credible}, ASTON only overlaps slightly with the approach of Camargo/random, proving that the results of ASTON are statistically different from those of the other tested approaches, except Camargo/random. Note that the statistical tests are performed without taking into account the event logs \textit{Sepsis} and \textit{Nasa} since the approaches of Camargo cannot be executed in those event logs.

\begin{table}[t!]
    \centering
    \caption{Plackett-Luce rankings and posterior probabilities of the tested approaches.}
    \begin{tabular}{l|ccccccc}
        & \rotatebox{90}{ASTON} & \rotatebox{90}{\shortstack[l]{Camargo/ \\ random}} & \rotatebox{90}{Evermann} & \rotatebox{90}{\shortstack[l]{Camargo/ \\ argmax}} & \rotatebox{90}{Taymouri} & \rotatebox{90}{Tax} & \rotatebox{90}{\shortstack[l]{Francesco- \\ marino}} \\  \hline
        Rank & 1.06 & 2.29 & 2.98 & 3.91 & 5.25 & 5.57 & 6.92 \\ 
        Prob. (\%) & 43.55 & 20.06 & 14.59 & 9.81 & 5.41 & 4.66 & 1.89 \\  \hline
    \end{tabular}
    \label{tab:plackett-luce}
\end{table}

To better quantify the differences between ASTON and the second and third-best approaches of the Plackett-Luce ranking ---Camargo/random and Evermann--- a hierarchical Bayesian test has been performed. Given a pair of approaches $A$ and $B$, this test obtains the probability that $A$ is better than $B$ ($A > B$), that $A$ has the same performance as $B$ ($A = B$), and that $A$ is worse than $B$ ($A < B$). Thus, the approach $A$ is better than $B$ if $A > B$ is statistically significant, and, equivalently, the approach $B$ is no worse than $A$ if the sum of $B > A$ and $A = B$ is statistically significant. In this test, it is considered that a probability greater than 95\% indicates statistical significance. The results of this hierarchical Bayesian test are shown in \tablename~\ref{tab:hierarchical-1} and \ref{tab:hierarchical-2}:

\begin{itemize}
    \item \tablename~\ref{tab:hierarchical-1} shows that ASTON significantly outperforms Camargo/random in five event logs ---\textit{BPI 2012 W}, \textit{BPI 2013 C.P.}, \textit{BPI 2013 I.}, \textit{Env. permit}, and \textit{Helpdesk}---, being close to be significantly better in \textit{BPI 2012 W C.} ---with 83.49\% of probability---. For the other five event logs, there is a high degree of uncertainty to make any conclusion: Camargo/random does not significantly outperform ASTON in any event log, being close in \textit{BPI 2012 A} and \textit{BPI 2012 C.} ---with 76.99\% and 80.51\% of probability, respectively---. These results suggest that ASTON outperforms Camargo/random in event logs that have shorter traces, while it does not underperform Camargo/random when traces are longer, showing that ASTON is overall more robust to the underlying characteristics of the event log.

\begin{table}[t!]
    \centering
    \caption{Hierarchical Bayesian tests per dataset: ASTON (A) vs Camargo/random (C).}
    \begin{tabular}{l|ccc}
         Event log & $A > C$ & $A = C$ & $A < C$ \\ \hline
         BPI 2012 & 61.51\% & 0.68\% & 37.81\% \\ 
         BPI 2012 A & 11.44\% & 11.56\% & 76.99\% \\
         BPI 2012 C. & 12.35\% & 7.14\% & 80.51\% \\
         BPI 2012 O & 39.51\% & 2.22\% & 58.27\% \\
         BPI 2012 W & 95.04\% & 0.98\% & 3.97\% \\
         BPI 2012 W C. & 83.49\% & 0.3\% & 16.20\% \\
         BPI 2013 C. P. & 97.58\% & 0.48\% & 1.93\% \\
         BPI 2013 I. & 99.90\% & 0.01\% & 0.08\% \\
         Env. permit & 99.48\% & 0.07\% & 0.44\% \\
         Helpdesk & 99.97\% & 0\% & 0.03\% \\ \hline
    \end{tabular}
    \label{tab:hierarchical-1}
\end{table}

    \item \tablename~\ref{tab:hierarchical-2} shows that ASTON significantly outperforms the Evermann approach for every event log, except for \textit{BPI 2012 W C.}, in which Evermann does not significantly outperforms ASTON ---having only the 58.28\% of probability---. Therefore, an encoder-decoder learning approach with an attention mechanism is an effective method for the activity suffix prediction due to less accumulation of errors during the prediction phase.
\end{itemize}

\begin{table}[t!]
    \centering
    \caption{Hierarchical Bayesian tests for ASTON (A) vs Evermann (E).}
    \begin{tabular}{l|ccc}
         Event log & $A > E$ & $A = E$ & $A < E$ \\ \hline
         BPI 2012 & 99.99\% & 0.01\% & 0\% \\ 
         BPI 2012 A & 99.90\% & 0\% & 0.1\% \\ 
         BPI 2012 C. & 99.99\% & 0.01\% & 0\% \\
         BPI 2012 O & 99.94\% & 0\% & 0.06\% \\
         BPI 2012 W & 99.90\% & 0\% & 0.1\% \\
         BPI 2012 W C. & 26.98\% & 5.01\% & 68.01\% \\
         BPI 2013 C. P. & 97.76\% & 0.46\% & 1.78\% \\
         BPI 2013 I. & 100\% & 0\% & 0\% \\
         Env. permit & 99.99\% & 0\% & 0.01\% \\
         Helpdesk & 99.99\% & 0\% & 0.01\% \\ \hline
    \end{tabular}
    \label{tab:hierarchical-2}
\end{table}

%%%%%%%%%%%%%%%%%%%%%%%%%%%%%%%%%%%%%%%%%%
\subsection{Inference strategy comparison}
%%%%%%%%%%%%%%%%%%%%%%%%%%%%%%%%%%%%%%%%%%

We have compared the performance of ASTON using different inference strategies to select the next activity during the suffix prediction phase. In particular, we compare the performance of ASTON by selecting the most likely activity ---\textit{argmax}---, randomly sampling over the next activity probability distribution ---\textit{random}---, using a beam search strategy ---\textit{beam}---, and normalizing the beam search log probabilities according to the length of the predicted suffix ---\textit{beam norm.}---. \tablename~\ref{tab:ablation-1} shows the performance of ASTON using these inference strategies.

On the one hand, note that the results reproduce the behavior highlighted in~\citep{rama-maneiro_deep_2022}, in which the \textit{argmax} strategy tends to perform better in logs with short traces such as \textit{BPI 2012 A}, \textit{BPI 2012 O}, \textit{\textit{BPI 2012 C.P.}}, \textit{Env. Permit} or \textit{Helpdesk} ---with an average difference of 2.58\%---, whereas the \textit{random} strategy works much better in event logs with longer traces such as \textit{BPI 2012}, \textit{BPI 2012 C.}, \textit{BPI 2012 W}, \textit{Nasa} or \textit{Sepsis} ---with an average difference of 9.21\%---. This happens because \textit{argmax} tends to get stuck predicting the same activity or set of activities when the traces are too long. On the other hand, \textit{random} is able to escape this kind of behavior by randomly switching to another activity.

\begin{table}[!b]
    \centering
    \caption{Performance of ASTON under several inference strategies.}
    \begin{tabular}{l|cccc}
    Event log & Argmax & Random & Beam & Beam norm. \\ \hline
    BPI 2012 & 0.1787 & 0.3449 & 0.3776 & \textbf{0.4004} \\
    BPI 2012 A & 0.5740 & 0.5689 & 0.6323 & \textbf{0.6340} \\
    BPI 2012 C. & 0.2458 & 0.3935 & 0.4247 & \textbf{0.4423} \\
    BPI 2012 O & 0.5936 & 0.5592 & 0.6079 & \textbf{0.6087} \\
    BPI 2012 W & 0.1931 & 0.3326 & 0.3296 & \textbf{0.3897} \\
    BPI 2012 W C. & 0.1463 & 0.2877 & 0.2918 & \textbf{0.3366} \\
    BPI 2013 C. P. & 0.6938 & 0.6892 & \textbf{0.7061} & 0.7042 \\
    Helpdesk & 0.8783 & 0.8436 & \textbf{0.8922} & 0.8920 \\
    Sepsis & 0.2858 & 0.2933 & 0.3086 & \textbf{0.3811} \\
    BPI 2013 I. & 0.5605 & 0.5509 & 0.6018 & \textbf{0.6069} \\
    Env. permit & 0.7622 & 0.7120 & 0.8260 & \textbf{0.8268} \\
    Nasa & 0.4297 & 0.4295 & 0.4341 & \textbf{0.4545} \\ \hline
    \end{tabular}
    \label{tab:ablation-1}
\end{table}

On the other hand, unnormalized beam search outperforms both the argmax and random sampling strategies for all event logs, except for \textit{BPI 2012 W}, proving that, regardless of trace length, beam search is more consistent than both strategies. Furthermore, except for \textit{BPI 2013 C. P.} and \textit{Helpdesk}, the normalized beam search obtains better results than the unnormalized beam search in every event log; particularly in \textit{BPI 2012}, \textit{BPI 2012 C.}, \textit{BPI 2012 W}, \textit{BPI 2012 W C.}, \textit{Sepsis}, and \textit{Nasa}, which have a fairly long average trace length and where the average difference between both approaches is 3.97\%. Therefore, it can be concluded that the other strategies tend to favor predicting long suffixes, whereas the unnormalized beam search is prone to predict short suffixes, although, unlike \textit{argmax}, it is robust enough in event logs with long traces to avoid the problem of getting stuck predicting the same activity over and over, being able to escape from suboptimal predictions.

\begin{table}[!b]
    \centering
    \caption{Average predicted suffix length for each inference strategy compared to the ground truth average suffix length.}
    \begin{tabular}{l|c|cccc}
    Event log & Truth & Argmax & Random & Beam & Beam norm. \\ \hline
    BPI 2012 & 19.44 & 141.28 & 12.68 & 4.81 & 14.84 \\
    BPI 2012 A & 2.21 & 2.72 & 2.17 & 1.34 & 1.37 \\
    BPI 2012 C. & 10.90 & 61.21 & 8.00 & 3.43 & 6.86 \\
    BPI 2012 O & 3.42 & 2.79 & 3.17 & 2.48 & 2.55 \\
    BPI 2012 W & 15.55 & 129.18 & 14.15 & 2.60 & 13.14 \\
    BPI 2012 W C. & 6.77 & 50.24 & 3.80 & 1.54 & 7.27 \\
    BPI 2013 C. P. & 2.86 & 2.37 & 2.46 & 2.27 & 2.32 \\
    Helpdesk & 1.97 & 1.80 & 2.02 & 1.69 & 1.70 \\
    Sepsis & 11.11 & 16.21 & 8.75 & 2.90 & 5.46 \\
    BPI 2013 I. & 7.18 & 12.19 & 5.88 & 2.99 & 3.57 \\
    Env. permit & 2.88 & 2.88 & 2.83 & 2.57 & 2.59 \\
    Nasa & 14.40 & 10.49 & 10.53 & 10.45 & 15.27 \\
    \hline
    \end{tabular}
    \label{tab:lengths}
\end{table}

In order to clarify these effects, \tablename~\ref{tab:lengths} shows the average length of the predicted suffixes for each inference strategy compared to the average length of the ground-truth suffixes. These results reveal that \textit{argmax} is unable to predict the end of the case in some event logs such as \textit{BPI 2012}, \textit{BPI 2012 C.}, \textit{BPI 2012 W}, or \textit{BPI 2012 W C.}, where the average predicted suffix length is orders of magnitude greater than the ground truth average suffix length. Furthermore, even though the length of suffixes predicted with the \textit{random} strategy is closer to the ground truth lengths than the unnormalized beam one, the results diverge from the real suffixes, as \tablename~\ref{tab:ablation-1} shows, since the random strategy does not consider the overall suffix probability. Finally, \tablename~\ref{tab:lengths} highlights that the unnormalized beam search usually predicts shorter suffixes than the normalized beam search strategy, meaning that it tends to predict the end of the case before it really occurs.

%%%%%%%%%%%%%%%%%%%%%%%%%%%%%%%%%%%%%
\section{Conclusions and future work}
\label{sec:conclusions}
%%%%%%%%%%%%%%%%%%%%%%%%%%%%%%%%%%%%%

This paper presents ASTON, a predictive monitoring approach that focuses on predicting the next activity suffix given a running case. Unlike most state-of-the-art approaches, ASTON addresses this problem using an encoder-decoder architecture with an attention mechanism. Thus, our model is trained to predict the suffix directly, accumulating fewer errors during inference and focusing only on predicting those elements that are relevant to the task at hand, which, in this paper, are the activities of the event log. In addition, our beam search strategy successfully balances the exploration and exploitation of candidate suffix selection during the post-training phase. Moreover, the proposed normalization for the beam search improves the performance in logs with long traces, while not degrading the performance in logs with shorter traces.

ASTON has been evaluated in 12 real-life event logs against 6 approaches from the state of the art. The results highlight that ASTON significantly outperforms all the state-of-the-art approaches but for Camargo/random, although ASTON still outperforms this approach in 5 of the event logs, whereas it is not significantly worse in the remaining 5 ---Camargo/random cannot be tested in 2 of the event logs since they do not have resources---. Furthermore, ASTON proves to be more robust to the underlying characteristics of the event log, performing consistently across a wide variety of process logs.

% ASTON has been evaluated in 12 real-life event logs against 6 other approaches from the state of the art. The results highlight that ASTON obtains better overall results in the whole set of tested event logs. Furthermore, our approach is more robust to the underlying characteristics of the event log, performing well in either event logs with short or long traces.

As future work, we intend to improve both the training and inference architectures. On the one hand, in the training phase, the aim is to experiment with different attention operations, to better capture long dependencies between the prefix activities, and to introduce an attention mechanism in the outputs of the decoder, in the same fashion that transformers do it. On the other hand, in the inference phase, we plan to study new sampling techniques for improving the selection of the best suffix activities.

% As future work, we intend to incorporate information about the attributes of the event logs, experiment with different attention operations between the encoder and the decoder, and test attention in the outputs of the decoder, in the same fashion that transformers do it.

%%%%%%%%%%%%%%%%%%%%%%%%%%
\section*{Acknowledgments}
%%%%%%%%%%%%%%%%%%%%%%%%%%

This work has received financial support from the Consellería de Educación, Universidade e Formación Profesional (accreditation 2019-2022 ED431G-2019/04), the European Regional Development Fund (ERDF), which acknowledges the CiTIUS - Centro Singular de Investigación en Tecnoloxías Intelixentes da Universidade de Santiago de Compostela as a Research Center of the Galician University System, and the Spanish Ministry of Science and Innovation (grants PDC2021-121072-C21 and PID2020-112623GB-I00). Efrén Rama-Maneiro is supported by the Spanish Ministry of Education, under the FPU national plan (FPU18/05687). Furthermore, the authors also wish to thank the supercomputer facilities provided by CESGA.

%%%%%%%%%%%%%%%%%%%%%%%%%%%%%%%%%%%
\bibliographystyle{elsarticle-harv}
\bibliography{main}

\begin{thebibliography}{41}
\expandafter\ifx\csname natexlab\endcsname\relax\def\natexlab#1{#1}\fi
\expandafter\ifx\csname url\endcsname\relax
  \def\url#1{\texttt{#1}}\fi
\expandafter\ifx\csname urlprefix\endcsname\relax\def\urlprefix{URL }\fi

\bibitem[{Agarwal et~al.(2022)Agarwal, Gupta, Sindhgatta, and
  Dechu}]{reinforcement_agarwal_2022}
Agarwal, P., Gupta, A., Sindhgatta, R., Dechu, S., 2022. Goal-oriented next
  best activity recommendation using reinforcement learning. CoRR
  abs/2205.03219.

\bibitem[{Arjovsky et~al.(2019)Arjovsky, Bottou, Gulrajani, and
  Lopez{-}Paz}]{risk_arjovsky_2019}
Arjovsky, M., Bottou, L., Gulrajani, I., Lopez{-}Paz, D., 2019. Invariant risk
  minimization. CoRR abs/1907.02893.

\bibitem[{Bahdanau et~al.(2015)Bahdanau, Cho, and
  Bengio}]{bahdanau_neural_2016}
Bahdanau, D., Cho, K., Bengio, Y., 2015. Neural machine translation by jointly
  learning to align and translate. In: 3rd International Conference on Learning
  Representations ({ICLR} 2015).

\bibitem[{Benavoli et~al.(2017)Benavoli, Corani, Demsar, and
  Zaffalon}]{benavoli_time_2017}
Benavoli, A., Corani, G., Demsar, J., Zaffalon, M., 2017. Time for a change:
  {A} tutorial for comparing multiple classifiers through bayesian analysis.
  Journal of Machine Learning Research 18, 77:1--77:36.

\bibitem[{Bengio et~al.(2015)Bengio, Vinyals, Jaitly, and
  Shazeer}]{bengio_scheduled_2015}
Bengio, S., Vinyals, O., Jaitly, N., Shazeer, N., 2015. Scheduled sampling for
  sequence prediction with recurrent neural networks. In: 28th Annual
  Conference on Neural Information Processing Systems ({NIPS} 2015). {ACM}, pp.
  1171--1179.

\bibitem[{Brauwers and Frasincar(2021)}]{attention_brauwers_2021}
Brauwers, G., Frasincar, F., 2021. A general survey on attention mechanisms in
  deep learning. IEEE Transactions on Knowledge and Data Engineering.

\bibitem[{Calvo et~al.(2018)Calvo, Ceberio, and Lozano}]{calvo_bayesian_2018}
Calvo, B., Ceberio, J., Lozano, J.~A., 2018. Bayesian inference for algorithm
  ranking analysis. In: Proceedings of the Genetic and Evolutionary Computation
  Conference Companion ({GECCO} 2018). {ACM}, pp. 324--325.

\bibitem[{Calvo and Santaf{\'{e}}(2016)}]{calvo_scmamp_2016}
Calvo, B., Santaf{\'{e}}, G., 2016. {scmamp}: {S}tatistical comparison of
  multiple algorithms in multiple problems. {The R Journal} 8~(1), 248--256.

\bibitem[{Camargo et~al.(2019)Camargo, Dumas, and
  Rojas}]{camargo_learning_2019}
Camargo, M., Dumas, M., Rojas, O.~G., 2019. Learning accurate {LSTM} models of
  business processes. In: 17th International Conference on Business Process
  Management ({BPM} 2019). Vol. 11675 of Lecture Notes in Computer Science.
  Springer, pp. 286--302.

\bibitem[{Cho et~al.(2014)Cho, van Merrienboer, G{\"{u}}l{\c{c}}ehre, Bahdanau,
  Bougares, Schwenk, and Bengio}]{Cho2014}
Cho, K., van Merrienboer, B., G{\"{u}}l{\c{c}}ehre, {\c{C}}., Bahdanau, D.,
  Bougares, F., Schwenk, H., Bengio, Y., 2014. Learning phrase representations
  using {RNN} encoder{\textendash}decoder for statistical machine translation.
  In: 2014 Conference on Empirical Methods in Natural Language Processing
  ({EMNLP} 2014). {ACL}, pp. 1724--1734.

\bibitem[{Dalmas et~al.(2021)Dalmas, Baranski, and
  Cortinovis}]{entity_dalmas_2021}
Dalmas, B., Baranski, F., Cortinovis, D., 2021. Predicting process activities
  and timestamps with entity-embeddings neural networks. In: 15th International
  Conference on Research Challenges in Information Science ({RCIS} 2021). Vol.
  415 of Lecture Notes in Business Information Processing. Springer, pp.
  393--408.

\bibitem[{de~Leoni et~al.(2016)de~Leoni, van~der Aalst, and Dees}]{Leoni2016}
de~Leoni, M., van~der Aalst, W.~M., Dees, M., 2016. A general process mining
  framework for correlating, predicting and clustering dynamic behavior based
  on event logs. Information Systems 56, 235--257.

\bibitem[{Evermann et~al.(2017)Evermann, Rehse, and Fettke}]{Evermann2017}
Evermann, J., Rehse, J., Fettke, P., 2017. Predicting process behaviour using
  deep learning. Decision Support Systems 100, 129--140.

\bibitem[{Francescomarino et~al.(2019)Francescomarino, Dumas, Maggi, and
  Teinemaa}]{Francescomarino2019}
Francescomarino, C.~D., Dumas, M., Maggi, F.~M., Teinemaa, I., 2019.
  Clustering-based predictive process monitoring. {IEEE} Transactions on
  Services Computing 12~(6), 896--909.

\bibitem[{Francescomarino and Ghidini(2022)}]{francescomarino_2022}
Francescomarino, C.~D., Ghidini, C., 2022. Predictive process monitoring. In:
  Process Mining Handbook. Vol. 448 of Lecture Notes in Business Information
  Processing. Springer, pp. 320--346.

\bibitem[{Francescomarino et~al.(2017)Francescomarino, Ghidini, Maggi,
  Petrucci, and Yeshchenko}]{Francescomarino2017}
Francescomarino, C.~D., Ghidini, C., Maggi, F.~M., Petrucci, G., Yeshchenko,
  A., 2017. An eye into the future: {L}everaging a-priori knowledge in
  predictive business process monitoring. In: 15th International Conference on
  Business Process Management ({BPM} 2017). Vol. 10445 of Lecture Notes in
  Computer Science. Springer, pp. 252--268.

\bibitem[{Gali et~al.(2019)Gali, Mariescu{-}Istodor, Hostettler, and
  Fr{\"{a}}nti}]{Gali2019}
Gali, N., Mariescu{-}Istodor, R., Hostettler, D., Fr{\"{a}}nti, P., 2019.
  Framework for syntactic string similarity measures. Expert Systems with
  Applications 129, 169--185.

\bibitem[{Goodfellow et~al.(2016)Goodfellow, Bengio, and
  Courville}]{goodfellow_deep_2016}
Goodfellow, I., Bengio, Y., Courville, A., 2016. Deep Learning. MIT Press.

\bibitem[{Jalayer et~al.(2020)Jalayer, Kahani, Beheshti, Pourmasoumi, and
  Motahari{-}Nezhad}]{attention_jalayer_2020}
Jalayer, A., Kahani, M., Beheshti, A., Pourmasoumi, A., Motahari{-}Nezhad,
  H.~R., 2020. Attention mechanism in predictive business process monitoring.
  In: 24th {IEEE} International Enterprise Distributed Object Computing
  Conference ({EDOC} 2020). {IEEE}, pp. 181--186.

\bibitem[{Ketyk{\'{o}} et~al.(2022)Ketyk{\'{o}}, Mannhardt, Hassani, and van
  Dongen}]{averages_ketyko_2022}
Ketyk{\'{o}}, I., Mannhardt, F., Hassani, M., van Dongen, B.~F., 2022. What
  averages do not tell: predicting real life processes with sequential deep
  learning. In: 37th {ACM/SIGAPP} Symposium on Applied Computing ({SAC} 2022).
  {ACM}, pp. 1128--1131.

\bibitem[{Khan et~al.(2021)Khan, Le, Do, Tran, Ghose, Dam, and
  Sindhgatta}]{mann_khan_2021}
Khan, A., Le, H., Do, K., Tran, T., Ghose, A., Dam, H.~K., Sindhgatta, R.,
  2021. Deepprocess: {S}upporting business process execution using a
  {MANN}-based recommender system. In: 19th International Conference on
  Service-Oriented Computing ({ICSOC} 2021). Vol. 13121 of Lecture Notes in
  Computer Science. Springer, pp. 19--33.

\bibitem[{Lee et~al.(2018)Lee, Parra, Munoz-Gama, and
  Sep{\'{u}}lveda}]{Lee2018}
Lee, W. L.~J., Parra, D., Munoz-Gama, J., Sep{\'{u}}lveda, M., 2018. Predicting
  process behavior meets factorization machines. Expert Systems with
  Applications 112, 87--98.

\bibitem[{Leontjeva et~al.(2015)Leontjeva, Conforti, Francescomarino, Dumas,
  and Maggi}]{Leontjeva2015}
Leontjeva, A., Conforti, R., Francescomarino, C.~D., Dumas, M., Maggi, F.~M.,
  2015. Complex symbolic sequence encodings for predictive monitoring of
  business processes. In: 13th International Conference on Business Process
  Management ({BPM} 2015). Vol. 9253 of Lecture Notes in Computer Science.
  Springer, pp. 297--313.

\bibitem[{Lin et~al.(2019)Lin, Wen, and Wang}]{mmpred_lin_2019}
Lin, L., Wen, L., Wang, J., 2019. {MM-Pred}: {A} deep predictive model for
  multi-attribute event sequence. In: 2019 {SIAM} International Conference on
  Data Mining ({SDM} 2019). {SIAM}, pp. 118--126.

\bibitem[{Mauro et~al.(2019)Mauro, Appice, and Basile}]{Mauro2019}
Mauro, N.~D., Appice, A., Basile, T. M.~A., 2019. Activity prediction of
  business process instances with inception {CNN} models. In: XVIIIth
  International Conference of the Italian Association for Artificial
  Intelligence ({AI*IA} 2019). Vol. 11946 of Lecture Notes in Computer Science.
  Springer, pp. 348--361.

\bibitem[{Mehdiyev et~al.(2020)Mehdiyev, Evermann, and Fettke}]{Mehdiyev2018}
Mehdiyev, N., Evermann, J., Fettke, P., 2020. A novel business process
  prediction model using a deep learning method. Business {\&} Information
  Systems Engineering 62~(2), 143--157.

\bibitem[{Pasquadibisceglie et~al.(2019)Pasquadibisceglie, Appice, Castellano,
  and Malerba}]{pasquadibisceglie_cnn_2019}
Pasquadibisceglie, V., Appice, A., Castellano, G., Malerba, D., 2019. Using
  convolutional neural networks for predictive process analytics. In: 2019
  International Conference on Process Mining ({ICPM} 2019). {IEEE}, pp.
  129--136.

\bibitem[{Rama-Maneiro et~al.(2022)Rama-Maneiro, Vidal, and
  Lama}]{rama-maneiro_deep_2022}
Rama-Maneiro, E., Vidal, J., Lama, M., 2022. Deep learning for predictive
  business process monitoring: {R}eview and benchmark. {IEEE} Transactions on
  Services Computing.

\bibitem[{Rama{-}Maneiro et~al.(2021)Rama{-}Maneiro, Vidal, and
  Lama}]{rama-maneiro_embedding_2021}
Rama{-}Maneiro, E., Vidal, J.~C., Lama, M., 2021. Embedding graph convolutional
  networks in recurrent neural networks for predictive monitoring. CoRR
  abs/2112.09641.

\bibitem[{Singh and Singh(2020)}]{Singh2020}
Singh, D., Singh, B., 2020. Investigating the impact of data normalization on
  classification performance. Applied Soft Computing 97~(Part {B}), 105524.

\bibitem[{Sun et~al.(2021)Sun, Ying, Yang, and Shen}]{remaining_sun_2021}
Sun, X., Ying, Y., Yang, S., Shen, H., 2021. Remaining activity sequence
  prediction for ongoing process instances. International Journal of Software
  Engineering and Knowledge Engineering 31~(11{\&}12), 1741--1760.

\bibitem[{Tax et~al.(2020)Tax, Teinemaa, and van
  Zelst}]{tax_interdisciplinary_2020}
Tax, N., Teinemaa, I., van Zelst, S.~J., 2020. An interdisciplinary comparison
  of sequence modeling methods for next-element prediction. Software and
  Systems Modeling 19~(6), 1345--1365.

\bibitem[{Tax et~al.(2017)Tax, Verenich, Rosa, and Dumas}]{Tax2017}
Tax, N., Verenich, I., Rosa, M.~L., Dumas, M., 2017. Predictive business
  process monitoring with {LSTM} neural networks. In: 29th International
  Conference on Advanced Information Systems Engineering ({CAiSE} 2017). Vol.
  10253 of Lecture Notes in Computer Science. Springer, pp. 477--492.

\bibitem[{Taymouri et~al.(2021)Taymouri, Rosa, and
  Erfani}]{adversarial_taymouri_2021}
Taymouri, F., Rosa, M.~L., Erfani, S.~M., 2021. A deep adversarial model for
  suffix and remaining time prediction of event sequences. In: 2021 {SIAM}
  International Conference on Data Mining ({SDM} 2021). {SIAM}, pp. 522--530.

\bibitem[{Taymouri et~al.(2020)Taymouri, Rosa, Erfani, Bozorgi, and
  Verenich}]{next_taymouri_2020}
Taymouri, F., Rosa, M.~L., Erfani, S.~M., Bozorgi, Z.~D., Verenich, I., 2020.
  Predictive business process monitoring via generative adversarial nets: The
  case of next event prediction. In: Business Process Management - 18th
  International Conference, {BPM} 2020, Seville, Spain, September 13-18, 2020,
  Proceedings. Vol. 12168 of Lecture Notes in Computer Science. Springer, pp.
  237--256.

\bibitem[{van~der Aalst(2016)}]{van_der_aalst_process_2016}
van~der Aalst, W. M.~P., 2016. {Process Mining: Data Science in Action}, 2nd
  Edition. Springer.

\bibitem[{Vaswani et~al.(2017)Vaswani, Shazeer, Parmar, Uszkoreit, Jones,
  Gomez, Kaiser, and Polosukhin}]{vaswani_attention_2017}
Vaswani, A., Shazeer, N., Parmar, N., Uszkoreit, J., Jones, L., Gomez, A.~N.,
  Kaiser, L., Polosukhin, I., 2017. Attention is all you need. In: 30th Annual
  Conference on Neural Information Processing Systems ({NIPS} 2017). {ACM}, pp.
  5998--6008.

\bibitem[{Venkateswaran et~al.(2021)Venkateswaran, Muthusamy, Isahagian, and
  Venkatasubramanian}]{robust_venkateswaran_2021}
Venkateswaran, P., Muthusamy, V., Isahagian, V., Venkatasubramanian, N., 2021.
  Robust and generalizable predictive models for business processes. In: 19th
  International Conference on Business Process Management ({BPM} 2021). Vol.
  12875 of Lecture Notes in Computer Science. Springer, pp. 105--122.

\bibitem[{Venugopal et~al.(2021)Venugopal, T{\"{o}}llich, Fairbank, and
  Scherp}]{Venugopal2021}
Venugopal, I., T{\"{o}}llich, J., Fairbank, M., Scherp, A., 2021. A comparison
  of deep-learning methods for analysing and predicting business processes. In:
  2021 International Joint Conference on Neural Networks ({IJCNN} 2021).
  {IEEE}, pp. 1--8.

\bibitem[{Weinzierl et~al.(2020)Weinzierl, Dunzer, Zilker, and
  Matzner}]{weinzierl_prescriptive_2020}
Weinzierl, S., Dunzer, S., Zilker, S., Matzner, M., 2020. Prescriptive business
  process monitoring for recommending next best actions. In: Business Process
  Management Forum 2020. Vol. 392 of Lecture Notes in Business Information
  Processing. Springer, pp. 193--209.

\bibitem[{Wu et~al.(2016)Wu, Wu, Schuster, Chen, Le, Norouzi, Macherey, Krikun,
  and et~al.}]{wu_googlenmt_2016}
Wu, Y., Wu, Y., Schuster, M., Chen, Z., Le, Q.~V., Norouzi, M., Macherey, W.,
  Krikun, M., et~al., 2016. Google's neural machine translation system:
  {B}ridging the gap between human and machine translation. CoRR
  abs/1609.08144.

\end{thebibliography}
%%%%%%%%%%%%%%%%%%%%%%%%%%%%%%%%%%%

\end{document}